\documentclass[pdflatex,sn-vancouver-num]{sn-jnl}% Vancouver Numbered Reference Style

\usepackage{graphicx}%
\usepackage{multirow}%
\usepackage{amsmath,amssymb,amsfonts}%
\usepackage{amsthm}%
\usepackage{mathrsfs}%
\usepackage[title]{appendix}%
\usepackage{xcolor}%
\usepackage{textcomp}%
\usepackage{manyfoot}%
\usepackage{booktabs}%
\usepackage{algorithm}%
\usepackage{algorithmicx}%
\usepackage{algpseudocode}%
\usepackage{listings}%
\usepackage{siunitx}
\usepackage{hyperref}
\usepackage[capitalise]{cleveref}

\theoremstyle{thmstyleone}%

\theoremstyle{thmstyletwo}%

\theoremstyle{thmstylethree}%

\begin{document}

\title[Article Title]{Article Title}

\title[PiGrand]{PiGRAND: Physics-informed Graph Neural Diffusion for Intelligent Additive Manufacturing}

\author*[1,2]{\fnm{Benjamin} \sur{Uhrich}}\email{uhrich@informatik.uni-leipzig.de}

\author[1,2]{\fnm{Tim} \sur{Häntschel}}

\author[1,2]{\fnm{Erhard} \sur{Rahm}}

\affil[1]{\orgdiv{Center for Scalable Data Analytics and Artificial Intelligence Dresden/Leipzig}, \country{Germany}}
\affil[2]{\orgdiv{Leipzig University}, \orgaddress{\city{Leipzig}, \country{Germany}}}

\abstract{A comprehensive understanding of heat transport is essential for optimizing various mechanical and engineering applications, including 3D printing. Recent advances in machine learning, combined with physics-based models, have enabled a powerful fusion of numerical methods and data-driven algorithms. This progress is driven by the availability of limited sensor data in various engineering and scientific domains, where the cost of data collection and the inaccessibility of certain measurements are high. 
To this end, we present PiGRAND, a Physics-informed graph neural diffusion framework. In order to reduce the computational complexity of graph learning, an efficient graph construction procedure was developed. Our approach is inspired by the explicit Euler and implicit Crank-Nicolson methods for modeling continuous heat transport, leveraging sub-learning models to secure the accurate diffusion across graph nodes. To enhance computational performance, our approach is combined with efficient transfer learning. We evaluate PiGRAND on thermal images from 3D printing, demonstrating significant improvements in prediction accuracy and computational performance compared to traditional graph neural diffusion (GRAND) and physics-informed neural networks (PINNs). These enhancements are attributed to the incorporation of physical principles derived from the theoretical study of partial differential equations (PDEs) into the learning model. The PiGRAND code is open-sourced on GitHub: https://github.com/bu32loxa/PiGRAND}


\maketitle

\section{Introduction}
\label{Introduction}

Heat transfer or transport plays a crucial role in engineering and natural sciences. Accurate modeling of heat transport in such processes remains a complex task due to the highly dynamic and non-linear nature, especially in real-world applications. An understanding of heat transport in the context of additive manufacturing (AM) is a key factor in achieving superior production quality. Although traditional techniques, such as the finite element method (FEM) and the finite volume method (FVM), offer high accuracy by solving PDEs, they often require significant computational resources and a considerable amount of preprocessing effort for discretisation. 
\cite{waqar2021fem,li2017heat, roy2018heat}.\\
PINNs effectively combine the solution of PDEs with data-driven learning by incorporating weighting mechanisms to balance the focus between data and physical laws during optimization. Furthermore, they facilitate mesh-free solutions \cite{Raissi.2019}. However, they may exhibit limitations in terms of scalability due to the curse of dimensionality. This results in high computational costs and slow convergence, which prove an impediment when dealing with data. In recent years, graph neural networks (GNNs) have emerged as a powerful tool for learning on graph-structured data and have already shown promise in physical sciences \cite{li2020fourier, li2020multipole}. Specifically, GRAND offers a framework for modeling processes where information diffuses across nodes over time \cite{Chamberlain.2021}.\\
We propose PiGRAND, a novel framework that extends GRAND to model continuous heat transport. This is achieved through the introduction of several key innovations, with the objectives of accelerating the learning process and improving accuracy and scalability. In order to establish the foundation for graph learning algorithms and to reduce the computational complexity of graph diffusion operations, we present an efficient graph construction method for transforming thermal images into graph-structured data representing physical objects. Furthermore, a novel connectivity model is suggested to more accurately represent the spatial relationships in high-dimensional data. \\
This model is employed to ensure accurate diffusion across nodes with different properties. To improve prediction accuracy, we adopt the concept of PINNs and introduce a series of loss terms based on fundamental physical principles related to heat transport. In addition, an intelligent dissipation model is introduced to regulate the energy transfer at the boundaries. The utilisation of transfer learning facilitates the application of pre-trained models and knowledge derived from related tasks, thereby reducing the requirement for costly retraining on new data and accelerating the learning process \cite{zhuang2020comprehensive}. To enhance computational performance, we propose the utilisation of efficient transfer learning. We evaluate PiGRAND on the application of thermal images generated during 3D printing processes. Our results demonstrate a significant improvement in prediction accuracy for heat transfer compared to representatives of traditional GRAND and PINNs. This improvement is largely attributed to our key innovations, which allow the model to capture the underlying heat transport process more effectively. The main contributions of our work are as follows:\\
\begin{itemize}
\item We propose an efficient graph construction method for transforming thermal image data into graph-structured data.\\
\item We present explicit Euler- and implicit Crank-Nicolson inspired GRAND.\\
\item We extend GRAND by developing two sub-learning models (connectivity and dissipation) and integrating physical principles of heat transport as regularization techniques for graph learning.\\
\item We show that computational performance can be improved by the use of efficient transfer learning.\\
\item We present a comprehensive evaluation demonstrating that our framework is capable of predicting heat transport in 3D-printing, outperforming traditional GRAND and PINNs.\\
\end{itemize}
The remainder of this paper is organized as follows. In Section 2, we review related work contributing to graph-based neural diffusion models and their applications to physical process simulations. In Section 3, we present the detailed methodology of PiGRAND and its components, including the data transformation, the sub-learning models and the integration of physical principles as regularization techniques. Section 4 describes our results in the application of 3D printing. Section 5 presents a comprehensive evaluation, including the dataset of thermal images, the metrics used for evaluation and the comparison with GRAND and PINNs. Finally, in Section 6, we discuss the results and conclude the paper with potential avenues for future research.

\section{Related Work}
\label{Related Work}
This section reviews prior work most relevant to PiGRAND for heat transport modeling.
\subsection{Heat Transfer Modelling}
\label{Numerical Analysis}
The field of heat transfer represents one of the most challenging and widely studied areas within the discipline of computational mechanics. In the context of 3D printing technologies, the appearance of heat conduction is characterised by nonlinear behaviour and is influenced by a multitude of parameters. Scientific Computing enables engineers to simulate and predict the behaviour of thermal systems across a range of scales, from the component level to that of large-scale infrastructure. This facilitates the design, optimisation and analysis of products and processes. The development of numerical approximation methods has a long history, with the most widely used being the FEM and FVM \cite{wahyudi2021application, versteeg2007introduction,leveque2002finite, kang1996finite, hsu2012finite}.\\
Mukherjee \textit{et al.} have demonstrated their effectiveness in modeling heat transfer and fluid flow for a variety of materials and process parameters, including stainless steel, titanium, and aluminum alloys \cite{Mukherjee.2018, Mukherjee.2018b}.\\
Extensions of these methods have incorporated multiphysics coupling, meshfree formulations, and particle-based descriptions to improve robustness and accuracy in the presence of evolving domains and localized heat sources \cite{Ansari.2022, zohdi2014additive, zohdi2014direct, ganeriwala2014multiphysics, wessels2018metal, wessels2019investigation, Li.2014, Liu.2019b}.
The impact of residual stresses on the mechanical properties of 3D-printed lattices was investigated by Ahmed \textit{et al.} \cite{Ahmed.2022}.
While these techniques provide detailed physical insight, their computational cost and limited scalability often restrict their applicability in scenarios requiring repeated simulations, large-scale parameter studies or near-real-time inference. To address these challenges, recent work has explored reduced-order and learning-based surrogate models for thermal prediction in AM. Graph-based and meshfree representations have gained particular attention, as they naturally encode spatial neighborhoods and local interactions while avoiding the constraints of structured meshes \cite{riensche2023thermal, lu2023convolution}. Such representations enable flexible discretizations of complex geometries and provide a foundation for data-driven models that can learn heat propagation patterns directly from simulation or experimental data. These developments motivate the use of graph-structured learning frameworks that preserve the locality and physical interpretability of numerical heat conduction models while significantly reducing computational overhead. In this context, diffusion-based GNNs offer a promising avenue by aligning message-passing operations with discretized heat transport dynamics. The present work builds on this perspective by formulating a PiGRAND model specifically tailored to thermal modeling in AM.
\subsection{Physics-informed Learning}
Physics-informed learning has emerged as a powerful tool for incorporating prior physical knowledge into data-driven models, particularly for systems governed by PDEs. Raissi \textit{et al.} introduces PINNs as prominent class of such approaches, which embed governing equations, boundary conditions and constitutive relations directly into the loss function. This formulation enables neural networks to approximate solutions to forward and inverse PDE problems while enforcing physical consistency \cite{Raissi.2019}. Since their introduction, PINNs have been successfully applied to a variety of heat transfer and fluid flow problems. Wessels group developed the neural particle method for computational fluid dynamics and employed PINNs for continuum micromechanics \cite{henkes2022physics, wessels2020neural}.
Several studies have demonstrated the potential of physics-informed learning for thermal modeling in AM, for example by integrating conductive and convective heat transfer equations into neural network training objectives or by leveraging measurement data for real-time monitoring and anomaly detection \cite{Zhu.2021, zobeiry2021physics, Cai.2021, uhrich2024physics, Uhrich.2023}. 
In addition, a multi-model neural network approach was developed for condition monitoring \cite{Bauer.2023}. Xu \textit{et al.} employed a transfer learning approach based on PINNs for solving inverse problems in engineering structures under different loading scenarios \cite{xu2023transfer}. Rasht \textit{et al.} employed PINNs to solve acoustic wave propagation and full waveform inversion problems, demonstrating their meshless flexibility and strong inversion performance across varying structural complexities. \cite{rasht2022physics}. Hu \textit{et al.} introduced stochastic dimension gradient descent, a novel training methodology for scaling PINNs to solve extremely high-dimensional PDEs efficiently \cite{hu2024tackling}. Guo \textit {et al.} proposed a data-free predictive surrogate modeling framework, which employs tensor-decomposed convolutional neural networks to solve high-dimensional parametric problems without training data, achieving remarkable computational and memory efficiency on ultra large-scale simulations \cite{guo2025tensor}.
These approaches highlight the value of incorporating physical constraints to improve generalization under limited data and to reduce reliance on purely data-driven learning. 
In the present work, we build on the conceptual foundations of physics-informed learning while addressing its scalability limitations through a GRAND formulation that integrates physical regularization in a structurally consistent and computationally efficient manner.
\subsection{Differential-Equation-inspired Neural Architectures}
\label{Differential Equation-inspired Neural Networks}
Differential-equation–inspired neural architectures have emerged as an effective means of introducing physical structure and interpretability into deep learning models. Rather than treating neural networks as static input–output mappings, this paradigm views learning as the evolution of a dynamical system, where network depth corresponds to time discretization and layer updates mirror numerical integration schemes for ordinary or PDEs. This perspective provides a principled foundation for improving stability, robustness and generalization in deep models.
 Weinan established a formal connection between residual neural networks and forward Euler discretizations of dynamical systems, motivating the design of architectures inspired by classical time-integration methods \cite{He2016, E.2017}.
Shen \textit{et al.} extended this idea by incorporating a backward Euler formulations as implicit scheme, to enhance stability and allow for deeper networks without degradation\cite{Shen.2020}. He \textit{et al.} employed the use of ODE-inspired network design for single image super-resolution. The authors propose several network architectures based on Runge-Kutta methods \cite{He.2019}. Similar principles have also been adopted in ODE and PDE-inspired architectures, where the structure of motion, diffusion, transport or reaction equations informs the design of convolutional and recurrent neural networks \cite{Ruthotto.2020,park2023convolution, guo2025interpolating, Alt.2023, Uhrich.2023b}. Khoshsirat and Kambhamettu developed an ODE transformer network \cite{Khoshsirat.2023}.
A key advantage of differential-equation–inspired models is their ability to encode inductive biases that align learning dynamics with known physical processes. By borrowing concepts such as stability conditions, consistency and discretization error from numerical analysis, these architectures offer greater interpretability and improved training behavior compared to purely data-driven designs. This is particularly relevant for physical systems characterized by diffusive dynamics, where information propagation is inherently local and governed by conservation principles.\\
These ideas form the conceptual basis for diffusion-based learning on graphs, in which message passing can be interpreted as a discrete approximation of continuous diffusion processes over irregular domains. By extending differential-equation–inspired architectures to graph-structured data, GRAND models provide a natural and physically meaningful framework for modeling heat transport in complex geometries. The present work leverages this foundation by adopting both explicit and implicit diffusion schemes within a graph neural network to achieve stable and scalable thermal predictions.  
\subsection{Graph Neural Networks}
GNNs are designed to operate on graph-structured data, where entities are represented as nodes and interactions are encoded through edges. Unlike traditional neural networks that work with grid-like data such as images or sequences, GNNs are tailored to capture the complex, non-Euclidean structures found in social networks, molecular graphs and knowledge graphs \cite{li2023survey, fan2019graph, wang2022molecular, ye2022comprehensive, park2019estimating}. This formulation is particularly well suited for physical systems, as graphs naturally reflect spatial discretizations, neighborhood interactions and irregular geometries commonly encountered in scientific and engineering applications. By aggregating information from local neighborhoods, GNNs enable the modeling of complex dependencies that are difficult to capture with grid-based architectures.
In recent years, GNNs have been increasingly applied to physics-based problems, including particle interactions, fluid dynamics and surrogate modeling of PDEs. Schlomi \textit{et al.} employed a range of applications of GNNs within the context of particle physics \cite{shlomi2020graph}. In these contexts, message-passing mechanisms can be interpreted as localized information exchange analogous to numerical stencils, making GNNs a flexible alternative to traditional mesh-based solvers. Gao \textit{et al.} developed a novel discrete PINN framework based on graph convolutional network and variational structure of PDEs to solve forward and inverse PDEs \cite{gao2022physics}.
Despite their flexibility, conventional GNN architectures typically rely on fixed message-passing rules and shallow propagation depths, which can limit their ability to represent continuous diffusion processes over extended spatial or temporal scales. In physical systems governed by heat transport, this can lead to oversmoothing, numerical instability or insufficient representation of long-range thermal interactions. These limitations highlight the need for graph-based models that explicitly incorporate diffusion dynamics into their architectural design.
\subsection{Graph Neural Diffusion}
\label{Neural Graph Diffusion}
Chamberlain \textit{et al.} introduces GRAND models extending standard GNNs by interpreting message passing as a discretized diffusion process on a graph.
This perspective establishes a direct connection between graph learning and the numerical solution of PDEs, enabling the systematic design of stable and interpretable architectures. Atwood and Towsley introduced diffusion-based graph convolutions, while more recent approaches have formalized GNNs as explicit or implicit time discretizations of an underlying diffusion equation \cite{atwood2016diffusion}.  
A key contribution of diffusion-based graph models is their ability to mitigate common challenges in deep graph learning, such as oversmoothing and vanishing gradients, by leveraging well-established numerical integration schemes. Explicit formulations offer computational efficiency and simplicity, whereas implicit schemes provide improved stability and allow for deeper propagation without loss of expressive power \cite{Chamberlain.2021}. These properties have led to strong performance across a range of graph learning benchmarks and have enabled applications in domains such as climate modeling and environmental prediction \cite{Choi.2023,Jia.2023}.
Building on this foundation,  Thorpe \textit{et al.} have incorporated source terms and nonlinear dynamics to further enhance expressiveness \cite{thorpe2022grand++}. However, existing GRAND models are largely developed and evaluated on benchmark datasets, missing the integration of domain-specific physical constraints. In contrast, the present work introduces PiGRAND, a physics-informed graph neural diffusion framework tailored to heat transport modeling in AM. By embedding physically motivated regularization terms derived from heat conduction theory and introducing sub-learning models for connectivity and dissipation, PiGRAND extends diffusion-based graph learning. This work represents a significant extension of the ideas initially set forth in \cite{uhrich2024neural}. The conference paper introduces the explicit inspired GRAND model but lacks any evaluation and only provides preliminary results.
In contrast, the proposed work significantly extends the findings by presenting the implicit Crank-Nicolson-inspired GRAND model, incorporating a transfer learning strategy based on a foundation model to predict heat transport on other components based on different material with much greater efficiency. Additionally, it offers a comprehensive evaluation of the results, demonstrating the advantages compared to PINNs and traditional GRAND. The initial model from \cite{uhrich2024neural}, was also used in further work to generate temperature features as part of a multimodal graph transformer approach \cite{uhrich2025mpgt}.
\section{Methodology}
Our proposed framework features a graph construction method that transforms thermal data into graph-structured data, and our PiGRAND model. The theoretical background of these methods is given below.
\subsection{Data Transformation}
A graph $G=(V,E)$ is a data structure consisting of a set of vertices $V=\{1,\hdots ,N\}$ and a set of edges $E \subseteq V\times V$. If the graph is embedded in Euclidean space, i.e. there is a map $\iota: V\rightarrow \mathbb R^n$, the Euclidean distance of vertex positions $\iota(i) \in \mathbb{R}^n,\ i=1,\dots,N$ gives rise to a distance function on $V$. As a shorthand notation, we write $v_i:= \iota(i)$. We consider physical objects and therefore always have $n=3$. An arbitrary spatial structure in $\mathbb R^3$ can be approximated using a simplicial $3$-complex. For each layer in a printing job, the objective is to represent the partial object printed up to the respective layer by a simplicial complex, such that a) the shape of the complex closely resembles the shape of the part, and b) the dynamics of the heat distribution in the part can be modeled by a diffusion process on the underlying graph, i.e. the graph whose vertices and edges are given by the $0$- and $1$-simplices in the complex.
Thermal images of the surface are taken periodically during the printing process, allowing for detecting the shape of each layer by taking an empirical threshold. Stacking this information for all layers, we obtain a 3-dimensional set of pixels, representing the shape of the object.
\subsubsection{Graph Construction from Thermal Images}
For a detailed description of the underlying sensor data and thermal images, see \cite{uhrich2024physics}.  
We associate the pixels of a thermal image with points on a plane in $\mathbb{R}^3$, that is parallel to the (x, y)-plane and contains the current layer of the printing job. A point is assumed to be part of the printed object, if the temperature value of the associated pixel is above an empirical threshold. If this is the case, the point is added to a point cloud in $\mathbb{R}^3$, by attaching a third component to the pixel coordinates, encoding the vertical position of the layer. The empirical threshold is set to 423.15 Kelvin, which is approximately the temperature of the built plate and the unmelted metal powder. In order to build the simplicial complex, a representative subset needs to be selected from the composite point cloud. To this end, we make use of the pruning method described in \cite{kurz2021geometry}, which is based on iteratively removing the point with the highest scale-invariant density (SID).
\begin{figure}[ht]
 \includegraphics[width=.9\textwidth]{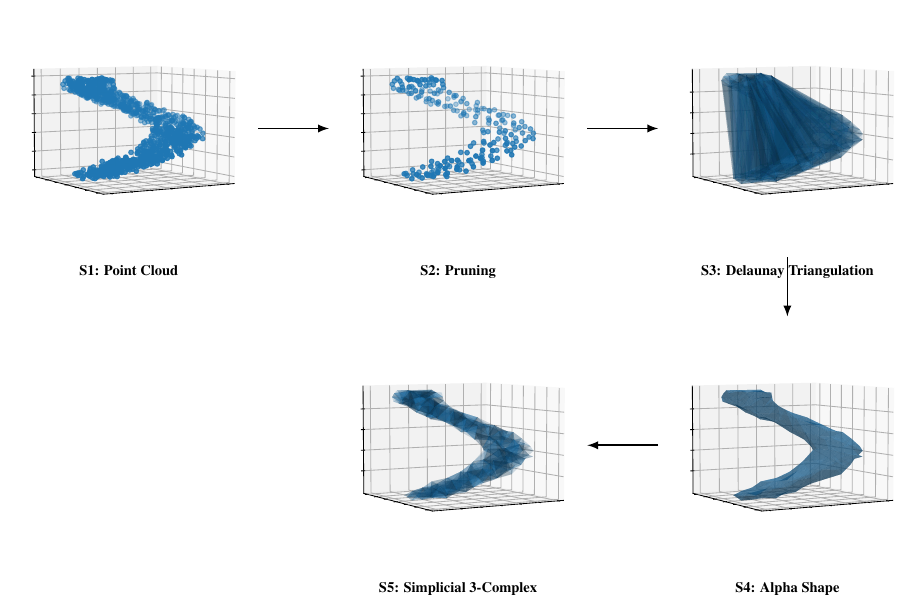}
 \caption{\textbf{Graph Construction Algorithm} - Step 1: Generating point cloud, Step 2: Pruning method, Step 3: Delaunay triangulation, Step 4: Alpha shape, Step 5: Simplicial 3-complex.}
\label{fig:data_integration}
\end{figure}
Since our point cloud is a set in $\mathbb{R}^3$ instead of a plane, we adapt the SID, by replacing the $r$-density with the 3-dimensional analogue
\begin{align}
    d_r(v_i) &= \frac{\#\{\Vert v_j-v_i\Vert_2 < r\ :\ 1\leq j \leq N,\ i\neq j\}}{\frac{4}{3} \pi r^3}, \\
    \intertext{which is the number of data points in the $r$-ball around $v_i$, divided by the volume of the ball. The scale-invariant density is defined as the integral over all $r$-densities:}
    d(v_i) &= \int_0^\infty d_r(v_i) dr\\
    \intertext{and by a similar calculation as in \cite{kurz2021geometry}, we see that:}
    d(v_i) &= \Big(\frac{8}{3} \pi \Big)^{-1} \sum_{j\neq i} \big\Vert v_i - v_j\big\Vert_2^{-2}.
\end{align}
In each step, the point with the highest spatial redundancy (measured by the SID) is removed. The resulting subset is relatively homogeneously distributed over the interior of the point cloud, but contains many boundary points, as their surrounding is partially void, resulting in a lower SID. This is desirable, as the boundary points define the shape of the printed object. 
For building the graphs representing the partially printed object, we iterate over the number of layers $n$, ranging from $1$ to the total number $M$ of layers in the printing job. For each $n$, the pruned set of points from the previous step is considered together with the points from the $n$-th layer. \\
We restrict pruning to points from the top $k$ layers ($k \ll M$), thus the representation for the first $n - k$ layers is inherited. A simplicial complex is constructed from the set of pruned points, using \textit{Delaunay triangulation} \cite{chen2004optimal}. The printed part must not necessarily be convex, but the shape produced by the Delaunay triangulation always is. To extract only the simplices that are within the boundary of the printed part, we use an \textit{alpha shape} \cite{edelsbrunner1994three} to determine the hull of the point cloud. Removing the simplices that are not encased by the alpha shape, we end up with a simplicial 3-complex that resembles the shape of our printed object (see \cref{fig:data_integration}). The vertices and edges of this complex are used to define a graph. Moreover, the option exists to bypass the pruning method when creating a simplicial complex based on the original point cloud. However, there would be a trade-off in terms of computational complexity.\\
The vertices are categorized based on spatial position:
\begin{enumerate}
    \item vertices in the lowest layer are assigned to the \textbf{bottom boundary} class
    \item vertices in the surface layer are assigned to the \textbf{top boundary} class
    \item vertices that are part of a surface of the alpha shape, but neither in the top- nor bottom boundary are assigned to the \textbf{side boundary} class
    \item vertices that are not part of either of these classes are assigned to the \textbf{interior} class
\end{enumerate}
$C_i=C(v_i)$ denotes the class of the vertex $v_i,\ i=1,\hdots N$.

\subsection{Numerical Models for the Graph Diffusion Process}

Formally, the heat equation, which describes the heat diffusion process in a homogeneous body and the initial-boundary value problem is given by
\begin{align}
 \frac{\partial}{\partial t} T(x,t) = \alpha\, \Delta T(x,t),\,\,\, x\in \Omega,\,\,\, t\in [0,T] \label{eq:heat_eq} \\
T(x,t_0) = T_0, \,\, x\in \Omega \label{eq:initial} \\ 
 \frac{\partial}{\partial n} T(x,t) = T_{\mathcal{B}} , \,\, x\in \Omega,\,\, t \in [0,T]
 \label{eq: boundary_cond}
\end{align}
where $\frac{\partial}{\partial t}$ is the derivative w.r.t. time, $\alpha$ is a conductivity parameter and $\Delta$ is the Laplace operator in the spatial domain.
For computational modeling of the heat transfer process, we must discretise \cref{eq:heat_eq} both in space and in time. In the context of a diffusion process on an object that is represented by a graph, the natural replacement for the Laplacian operator is the graph Laplacian matrix $L$, which is defined as
\begin{align}
    L := D - A \label{eq:def_graph_laplacian}
\end{align}
where $A$ is the adjacency matrix containing the edge weights for pairs of vertices in a fixed enumeration, and $D$ is the degree matrix, i.e. the diagonal matrix whose entries are the sum of the weights of adjacent edges for each vertex.\\
Regarding the time derivative, the practical replacement is found by considering difference quotients instead of the differential. To this end, the time domain $\big[0,T\big]$ is subdivided into small intervals of some length $\delta t \ll T$, and the discrete evaluation points are chosen as $t_n = n\,\delta t,\ n=0,1,\dots, T/\delta t$. When approximating the time derivative at a lattice point $t_n$, one must decide between selecting the difference quotient w.r.t. the previous or w.r.t. the subsequent lattice point. Depending on this decision, the approximation of the heat transfer process is either described by an explicit scheme motivated by the Taylor series for a solution of the homogeneous heat equation \cref{eq:heat_eq}:
\begin{align}
    T(x, t+\delta t) &= \sum_{j=0}^\infty \frac{1}{j!} \left(\left(\frac{\partial}{\partial x}\right)^j T \right)\big(x,t\big) (\delta t)^j \\
    &= T(x,t) + \sum_{j=1}^\infty \frac{1}{j!} \left(\left(\frac{\partial}{\partial x}\right)^j T \right)\big(x,t\big) (\delta t)^j
\end{align}
Note that for a solution of the homogeneous heat equation, its derivative w.r.t. $t$ is again a solution of the heat equation, since $0 = \frac{\partial}{\partial t} \big(\dot{T}-\alpha \Delta T\big) = \ddot{T}-\alpha \Delta \dot{T}$. Hence, we can replace $\big(\frac{\partial}{\partial x}\big)^j$ by $(\alpha\Delta)^j$ in the Taylor series. Approximating $\alpha \Delta$ by $L(T)$, we obtain:
\begin{align}
    T(x,t+\delta t) &\approx T(x,t) +  \sum_{j=1}^\infty \frac{1}{j!} \left(L(T)^j T \right)\big(x,t\big) (\delta t)^j
\end{align}
As can be seen, using only the first two terms of the Taylor series:
    \begin{align}
        &&\frac{T_{n+1} - T_n}{\delta t} &= L T_{n} && \label{eq:explicit_init}\\
        &\Longleftrightarrow & T_{n+1} &= T_n + \delta t L T_n \label{eq:explicit_step}&&
    \intertext{for the forward step (evaluated at time-step $n$), or an implicit scheme}
        &&\frac{T_{n+1} - T_{n}}{\delta t} &= L T_{n+1} && \label{eq:implicit_init}\\
        &\Longleftrightarrow & \big(\text{Id}-\delta t L\big) T_{n+1} &= T_{n} && \\
        &\Longleftrightarrow &  T_{n+1} &= \big(\text{Id}-\delta t L\big)^{-1} T_{n} && 
    \intertext{when choosing the backward step (evaluated at time-step $n+1$) and  introducing the identity matrix Id. These methods for solving PDEs numerically are known as the forward- and the backward Euler method. The LHS in \cref{eq:explicit_init} and \cref{eq:implicit_init} can be interpreted as an estimate for the derivative at the midpoint between times $t_n$ and $t_{n+1}$. On the other hand, the RHSs are the Laplacians at the times $t_n$ and $t_{n+1}$. A third scheme can be obtained by combining the two previous schemes, taking the mean of the Laplacian at $t_n$ and at $t_{n+1}$, in order to estimate the Laplacian at the midpoint,}
        &&\frac{T_{n+1} - T_{n}}{\delta t} &= \frac{1}{2} \Big( L T_{n} + L T_{n+1} \Big) &&\\
        &\Longleftrightarrow & \Big(\text{Id}-\frac{1}{2}\delta t L\Big) T_{n+1} &= \Big(\text{Id}+\frac{1}{2}\delta t L\Big) T_{n}. &&\\
        &\Longleftrightarrow &  T_{n+1} &= \Big(\text{Id}-\frac{1}{2}\delta t L\Big)^{-1}\Big(\text{Id}+\frac{1}{2}\delta t L\Big) T_{n}. && \label{eq:cn_step}. 
    \end{align}
     This is known as the \textit{Crank-Nicolson} scheme. Since $L$ has non-positive spectrum, $\big(\text{Id}-\delta t L\big)^{-1}$ and $\big(\text{Id}-(\delta t/2) L\big)^{-1}$ exist $\forall\ \delta t > 0$, and thus the implicit step and the Crank-Nicolson step both have a unique solution. The explicit method is efficient to compute (note that $L$ is sparse and the number of neighbours of a node in a mesh is constant for different sizes of the mesh, so the cost of \cref{eq:explicit_step} is linear in the number of nodes), but it requires choosing a small step size $\delta t$ of the order $(\delta x)^2$, otherwise the numerical solutions may explode \cite{bartels2016numerical}. In contrast, the implicit scheme and the Crank-Nicolson scheme are numerically stable for any choice of step size. 

\subsection{Neural Diffusion Models on Graphs}
The utilisation of numerical mathematics is fundamental to the creation of graph learning models that are capable of discretising and approximately solving the continuous heat equation.
Due to non-equidistant vertices in the graph, inhomogeneous conductivity depending on the temperature and a random laser trajectory in the 3D printing process, it is practically impossible to determine the correct parameters and boundary conditions for modeling the heat transport process purely numerically. Instead we propose a more complex model to predict the heat transfer, incorporating real measurement data as well as known properties of diffusion processes.
Revisiting \cref{eq:heat_eq}, we introduce a local state-dependence to $\alpha$, and an additional dissipation term $Q$, to account for heat loss at the boundary:
\begin{align}
    \dot{T}(x,t) = \alpha \big(T(x,t)\big) \Delta T(x,t) - Q\big(x,T(x,t)\big)
\end{align}

For a discrete approximation of the heat process on a graph, we consider the explicit scheme \cref{eq:explicit_step} and the Crank-Nicolson scheme \cref{eq:cn_step}, and extend both by introducing the dissipation term $Q$ on the RHS, which depends on the temperature state $T_n$. Furthermore, in the neural diffusion model, we allow for temperature-dependent conductivity, which implies that the graph Laplacian $L$ is a function of the temperature state. 
We build these functional dependencies, such that only the local graph structure and local temperature values influence the respective entries of $L$ and $Q$. For a single vertex $v_i$, we assert that the local information is given by the temperature $T_n(v_i)$, the vertex class $C(v_i)$ and the scale invariant density $d(v_i)$. For an edge between two adjacent vertices $v_i,v_j$, we define the local information as the vertex distance $\varrho_{ij} = \Vert v_i-v_j\Vert_2$, together with the local information for each of the two vertices. The graph Laplacian is then constructed as given by \cref{eq:def_graph_laplacian}, but the adjacency matrix $A=\big(a_{ij}\big)_{i,j=1,\dots,N}$ is replaced by a state-dependent adjacency $\widehat A$, defined as
\begin{align}
    \widehat A = \begin{cases}
        0, & a_{ij} = 0,\\
        c_{ij}, & a_{ij} \neq 0
    \end{cases},    
\end{align}
where the non-zero entries $c_{ij}$ of $\widehat A$ are estimated by a learnable function
\begin{align}
 c_{ij}(T_n) = \varphi\big(\varrho_{ij},T_n(v_i),T_n(v_j),C(v_i),C(v_j),d(v_i), d(v_j)\big)   
\end{align}
which is realised by a single-hidden-layer neural network of width $256$. Finally, for assembling the Laplacian, the degree matrix $D$ of $A$ is replaced accordingly by $\widehat D$ which is computed w.r.t. to $\widehat A$, such that we obtain the state dependent Laplacian $\widehat L(T_n) := \widehat D(T_n) - \widehat A(T_n)$.\\
In a similar fashion, the $i$-th entry of the dissipation vector $Q(T_n) = \big(Q_i(T_n)\big)_i$ should only depend on the local properties of $v_i$, i.e. $T_n(v_i)$, $C(v_i)$, $d(v_i)$. This motivates modelling $Q_i(T_n)$ by a function
\begin{align}
    Q_i(T_n) = \psi\big(T_n(v_i),C(v_i),d(v_i)\big)
\end{align}
which again is implemented as a single-hidden-layer network of width $256$.\\
Thus, in the explicit neural heat model, the diffusion process is computed recursively by the model equation
\begin{align}
    T_{n+1}(T_n) = T_n + \delta t\, \widehat{L}(T_n)\, T_n - \delta t\, Q(T_n) \label{eq:explicit_model_equation}
\end{align}
and in the neural heat model based on the Crank-Nicolson method, the adapted model equation is given by
\begin{align}
     T_{n+1}(T_n) &= \Big(\text{Id}-\frac{1}{2}\delta t\, \widehat L(T_n)\Big)^{-1}\Big(\text{Id}+\frac{1}{2}\delta t\, \widehat L(T_n)\Big)\, T_{n}- \delta t\,Q(T_n).
\end{align}

These models contain the trainable submodels $\varphi$ and $\psi$ determining $\widehat L$ and $Q$.\vspace{4px}\\
Training this kind of model presents several obstacles. We identify the following challenges:\\ 
\begin{enumerate}
    \item Using the model to predict changes in surface temperature requires knowing the initial state $T_0$ for all vertices, including those that cannot be observed. The limited observability of vertices likely leads to an underdetermined optimization problem, which raises doubts about the model's ability to accurately represent the internal heat state.\\
    \item The recursive nature of the model, without the possibility to correct the intermediate temperature state due to limited information, facilitates the compounding of errors and hinders the training process.\\
    \item The local character of the model, where in each step only the state in the direct neighbourhood affects the temperature change at a vertex, leads to vanishing gradients for the temperature values at vertices that are not in proximity to the surface layer.\\
\end{enumerate}
To address these issues, we deviate from the standard data-driven training process for machine learning models in the following manner:\\
\begin{itemize}[label=$\bullet$]
    \item To overcome the problem of the unknown initial states, we start at the first layer, where the complete state can be observed. We train the diffusion model on the initial layer, until an acceptable accuracy is achieved. Then, we use the obtained model to predict the diffusion process for the now hidden vertices in the first layer, until the start of the printing process for the second layer. Now, we use the predicted internal state as the hidden state for predictions on the second layer, and update the hidden temperature values for each time step, according to the model prediction. Again, we train the model on the first and second layer, until acceptable precision is reached, before starting to train the model on the third layer. This method of gradually increasing the number of layers helps us obtaining consistent internal states, which are the foundation for precise model predictions.\\
    \item Besides the standard loss function that compares the model prediction to the observed data, we introduce additional regularizing loss functions, based on physical and mathematical information about the diffusion process. Models that incorporate equations describing physical systems as regularization functions in the training of neural networks are known as PINNs. This approach restricts our model by introducing an additional loss to dynamics that are inconsistent with real diffusion processes.\\
    \item In order to introduce a dependency between vertices with larger distance in the graph, we must increase the discrete time horizon in each training step. To this end, we subdivide each time step in our training data into multiple steps for the numerical approximation. Then, if one training step consists of $k$ steps of the numerical approximation, paths of length up to $k$ are considered in the training process and therefore vertices in the lower part of the graph can influence the state at the surface.\\
\end{itemize}
\subsection{Regularizing Loss Functions}
\begin{figure}
    \includegraphics[width=1.0\textwidth]{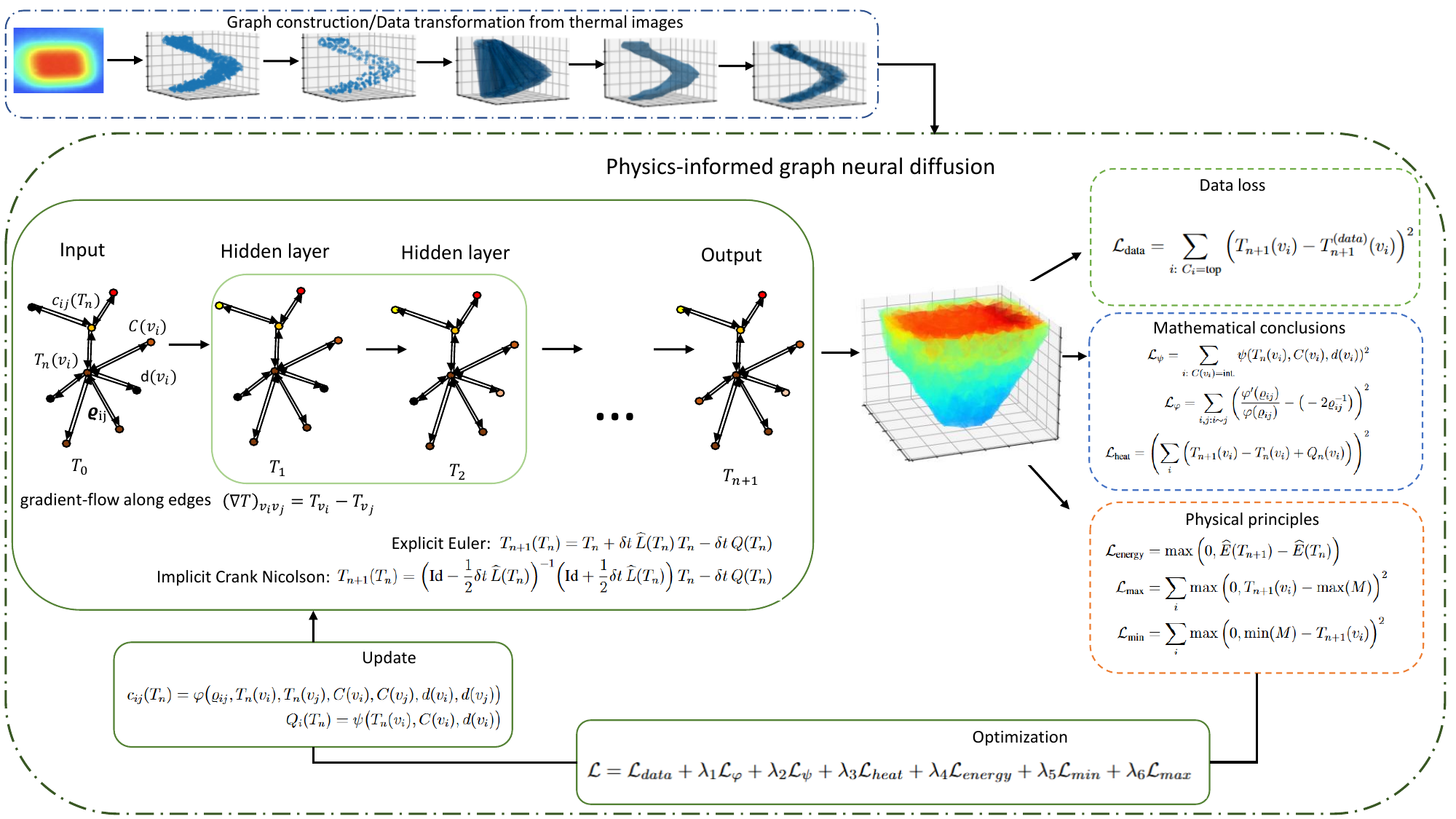}
    \caption{\textbf{Flowchart} - Graph construction and physics-informed graph neural diffusion}
    \label{fig:flowchart}
\end{figure}

We discuss in detail the derivation of the previously mentioned regularizing loss functions.\\
For the connectivity model $\varphi$, it is known from the discretization of the continuous Laplacian, that the connectivity of two adjacent vertices should be proportional to the inverse square of their distance, $\varrho_{ij}^{-2}$. Assuming $\varphi(\varrho_{ij}) = \tau \varrho_{ij}^{-2}$ (keeping all other parameters constant), it follows $\varphi'(\varrho_{ij}) = -2\tau \varrho_{ij}^{-3}$, and thus $\frac{\varphi'(\varrho_{ij})}{\varphi(\varrho_{ij})} = -2 \varrho_{ij}^{-1}$, which is independent of the scale $\tau$. Therefore, the first regularizing loss term is given by:
\begin{align}
    \mathcal{L}_{\varphi} = \sum_{i,j: i\sim j} \left(\frac{\varphi'(\varrho_{ij})}{\varphi(\varrho_{ij})} - \big(-2 \varrho_{ij}^{-1}\big)\right)^2
\end{align}
\begin{align}
\mathcal{L}_{\varphi} = \sum_{i,j: i\sim j} \left( \varphi (\varrho_{ij}) - \frac{1}{\varrho_{ij}^2} \right)^2
\end{align}
which ensures consistency of the edge weights with the distances of the connected vertices.
Furthermore, dissipation can only occur at the boundary, so $\psi=0$ is required for interior points, motivating the loss:
\begin{align}
    \mathcal{L}_{\psi} = \sum_{i:\ C(v_i)=\text{int.}} \psi(T_{n}(v_i),C(v_i),d(v_i))^2
\end{align}
Using knowledge from the theoretical study of PDEs, it is also possible to make statements about the temporal evolution of the heat state. First, the total thermal energy in the body can only change because of heat transfer at the boundary, which in the discrete model has the equivalent $\sum_i T_{n+1}(v_i) = \sum_i \Big(T_{n}(v_i) - Q_{n}(v_i)\Big)$. Therefore, we propose the loss function
\begin{align}
    \mathcal{L}_{\text{heat}} = \left(\sum_i \Big(T_{n+1}(v_i) - T_{n}(v_i) + Q_{n}(v_i)\Big)\right)^2.
    \label{eq:heat}
\end{align}
Another well known property of the evolution of heat distribution is the maximum principle (see for example \cite{evans2022partial}, §2.3.). For our purpose, it suggests that the temperature at a vertex is within the range given by the minimum and maximum over its previous temperature and the temperatures of connected vertices. This is expressed by the regularizing loss terms:
\begin{align}
    \mathcal{L}_{\text{max}} &= \sum_{i} \max\Big(0,T_{n+1}(v_i)-\max(M)\Big)^2
    \label{eq:max}
\end{align}

\begin{align}
    \mathcal{L}_{\text{min}} &= \sum_{i} \max\Big(0,\min(M)-T_{n+1}(v_i)\Big)^2\\
    M &= \big\{T_n(v_i);\ T_{n+1}(v_j),\ v_j\sim v_i\big\}\nonumber
    \label{eq:min}
\end{align}
Furthermore, a potential energy for the heat distribution can be defined by
\begin{align}
    E(T,t) =& \int_U \Big(T(x,t)-\overline{T}(t)\Big)^2 dx
\end{align}
For the time-differential of this energy, one can compute:
\begin{align}
    \dot{E}(T,t) =& 2\int_U \Big(T(x,t)-\overline{T}(t)\Big)\dot{T}(x,t) dx\nonumber \\
    =& 2\alpha \int_U \Big(T(x,t)-\overline{T}(t)\Big) \Delta T(x,t) dx\nonumber\\
    =&  2\alpha\bigg(\underbrace{\int_{\partial U} \Big(T(x,t)-\overline{T}(t)\Big) \big(\nabla T(x,t) \cdot \nu\big) dS}_{\text{energy loss from dissipation}} - \underbrace{\int_U |\nabla T(x,t)|^2 dx}_{\geq 0}\bigg)
\end{align}
Assuming dissipation is relatively small, it should roughly hold that $\dot{E}\leq 0$, so $E(T_{n+1}) \leq E(T_n)$. Thus, we introduce another loss term:
\begin{align}
    \mathcal{L}_{\text{energy}} &= \max\Big(0,\widehat{E}(T_{n+1})-\widehat{E}(T_n)\Big)
    \label{eq:energy}
\end{align}

where 
\begin{align}
    \widehat{E}(T_{n}) = \frac{1}{N}\sum_{i} \Big(T_n(v_i) - \overline{T}_n\Big)^2\ \text{ with }\ \overline{T}_n = \frac{1}{N}\sum_{i} T_n(v_i)
\end{align}
is the discrete approximation of the potential energy.\\
For training the model, the regularizing loss functions $\mathcal{L}_\varphi,\mathcal{L}_\psi,\mathcal{L}_{\text{heat}},\mathcal{L}_{\text{min}},\mathcal{L}_{\text{max}},$ and $\mathcal{L}_{\text{energy}}$, as well as the prediction loss
\begin{align}
    \mathcal{L}_{\text{data}} = \sum_{i:\ C_i=\text{top}} \Big(T_{n+1}(v_i)-T_{n+1}^{(data)}(v_i)\Big)^2
    \label{eq: data loss}
\end{align}
are added using appropriate weights, and the descent algorithm seeks to find a minimum for the sum, i.e. a model state that fits the data without violating known properties of heat diffusion. This restricts the admissible set of solutions, thus mitigating the problem of underdetermination, and the regularizing loss functions also regard vertices that are disconnected from the surface and therefore not captured by $\mathcal{L}_{\text{data}}$. A flowchart of the complete framework, including graph construction and PiGRAND, can be found in \cref{fig:flowchart}.
\begin{figure}[h!]
    \includegraphics[width=.9\textwidth]{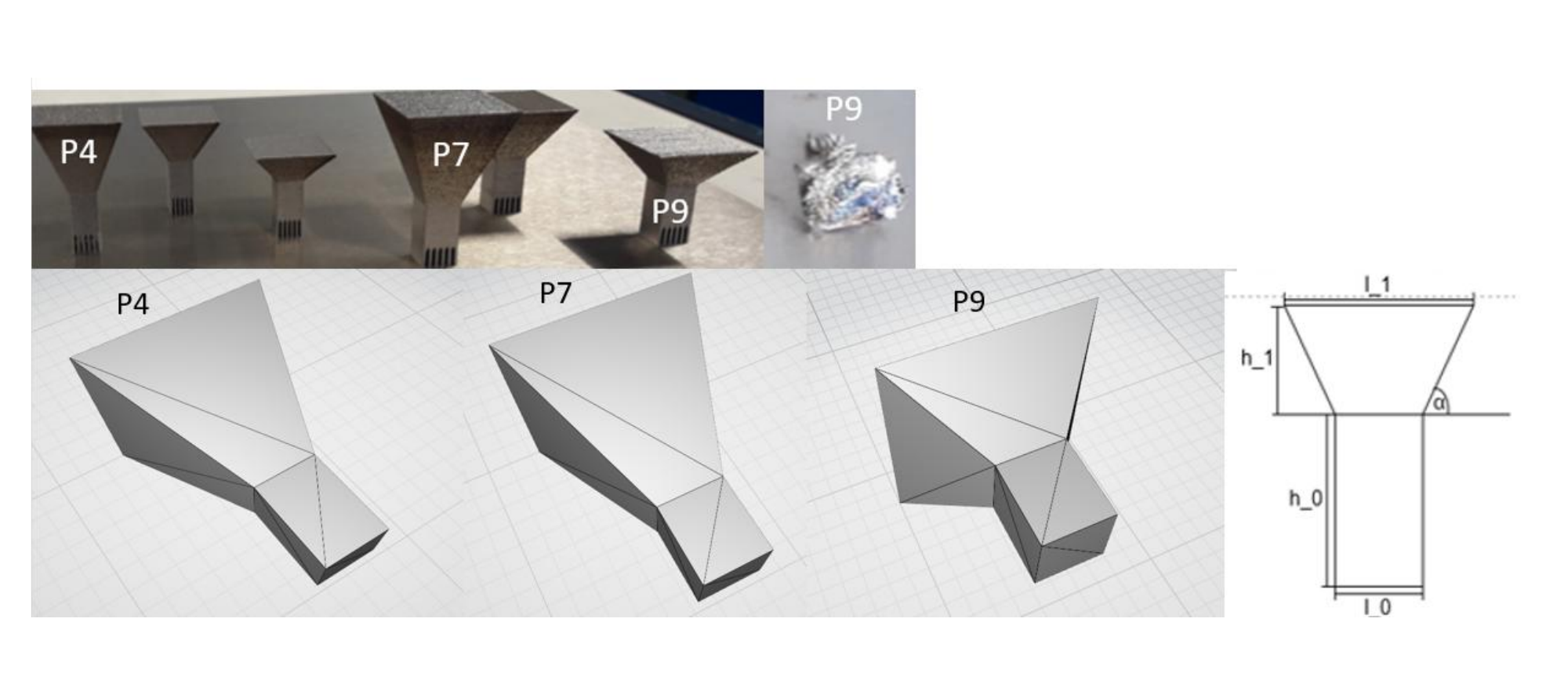}
    \caption{\textbf{Geometry of Printed Objects} - Photographs of high quality components, strongly deformed component P9 and CAD models}
    \label{fig:real-components}
\end{figure}
\begin{table}[h!]
    \centering
    \begin{tabular}{|c|c|c|c|c|c|c|}\hline
       P & $h_0$[mm] & $\alpha$ [°]& $l_1$[mm]&$l_0$[mm]& $h_1$[mm] & Material \\\hline
         4 & 10 & 70& 20&6.7& 18.27 & Stainless Steel\\\hline
        7 & 10 & 70& 25&8.3& 22.94 & Stainless Steel\\\hline
        9 & 10 & 50& 25&8.3& 9.95&  Stainless Steel\\\hline
        7M &  10 & 70 & 25 & 8.3 & 22.94 & Nickel-based Alloy \\\hline
    \end{tabular}
    \caption{Comparison of the measures of the 3D-printed objects P4, P7, P9 and P7M}
    \label{tab:measure}
\end{table}
\section{Results in the Application of Powder Bed Fusion}
The following section presents the results of PiGRAND and offers a comprehensive evaluation utilising a range of metrics. This analysis is designed to assess the performance, accuracy and overall effectiveness of the proposed method. Additionally, comparisons with state-of-the-art methods will be provided to highlight the relative strengths and potential areas for improvement in PiGRAND. This evaluation will provide insights into the robustness, scalability and practical applicability of the model across different datasets and scenarios. These include three printed stainless steel pyramids: two of good quality and one of poor quality and one of good quality made from a nickel-based alloy (see \cref{fig:real-components} and \cref{tab:measure}).
\subsection{Heat Transfer Prediction based on Thermal Images}
\begin{figure}
    \includegraphics[width=.9\textwidth]{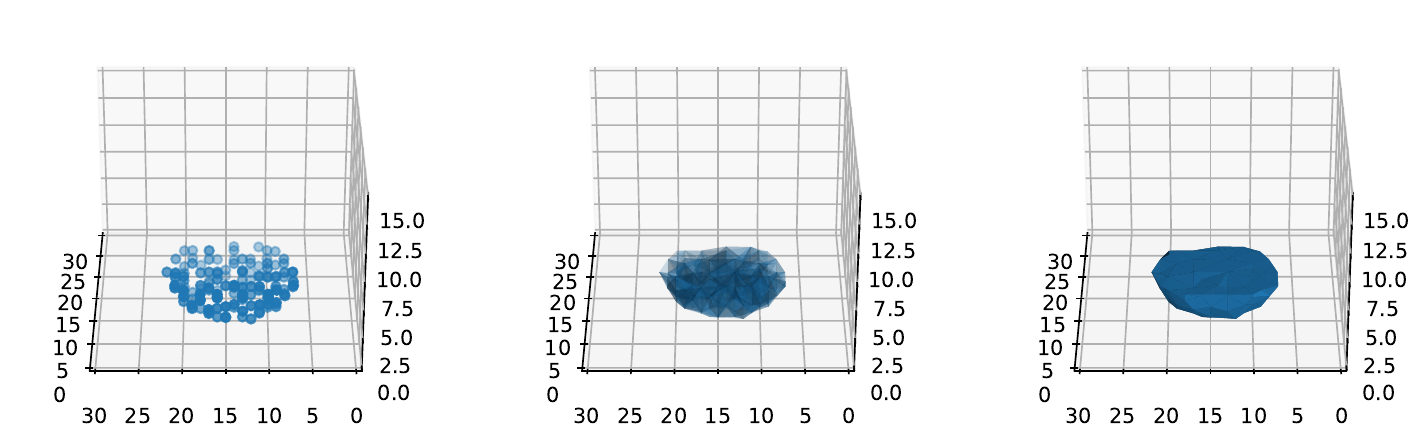}
    \includegraphics[width=.9\textwidth]{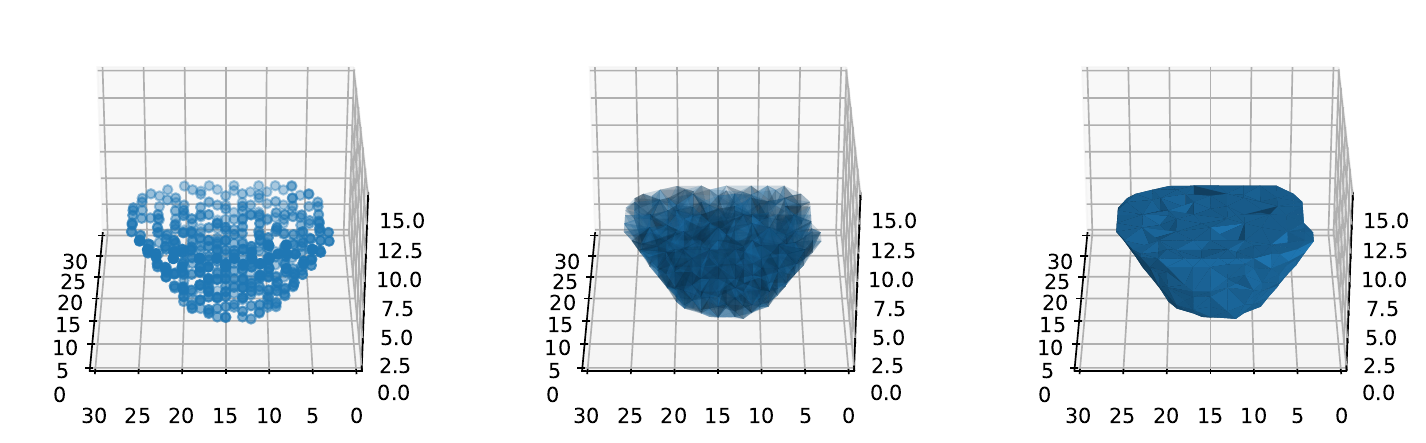}
    \includegraphics[width=.9\textwidth]{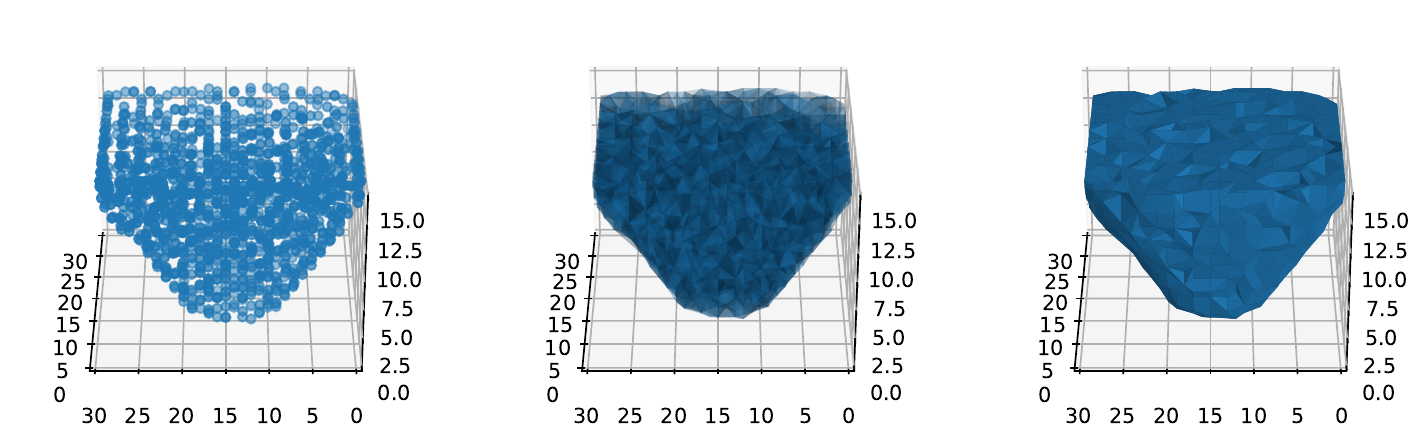}
    \caption{\textbf{Temporal-Spatial Graph} - graph model development for layer 100, 250, 500 (rows). Left column: point cloud, middle column: Simplicial Complex, right column: Alpha Shape}
    \label{fig:temporal graph}
\end{figure}

\begin{figure}
    \includegraphics[width=.5\textwidth]{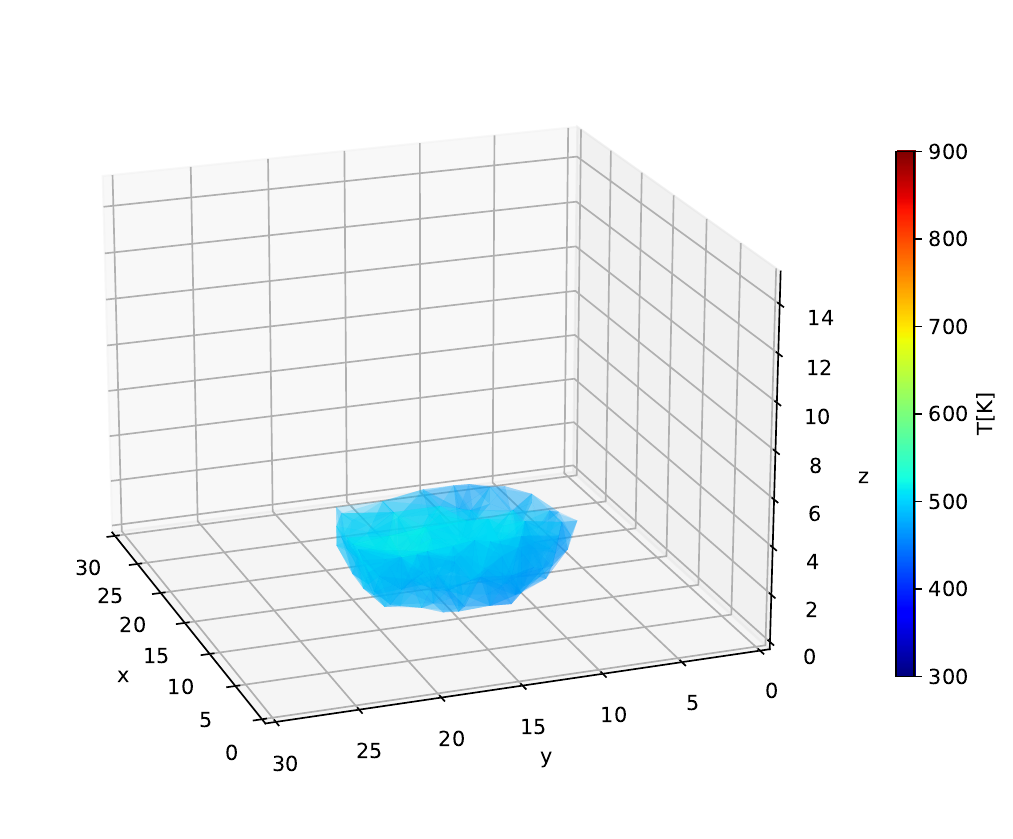}
    \includegraphics[width=.5\textwidth]{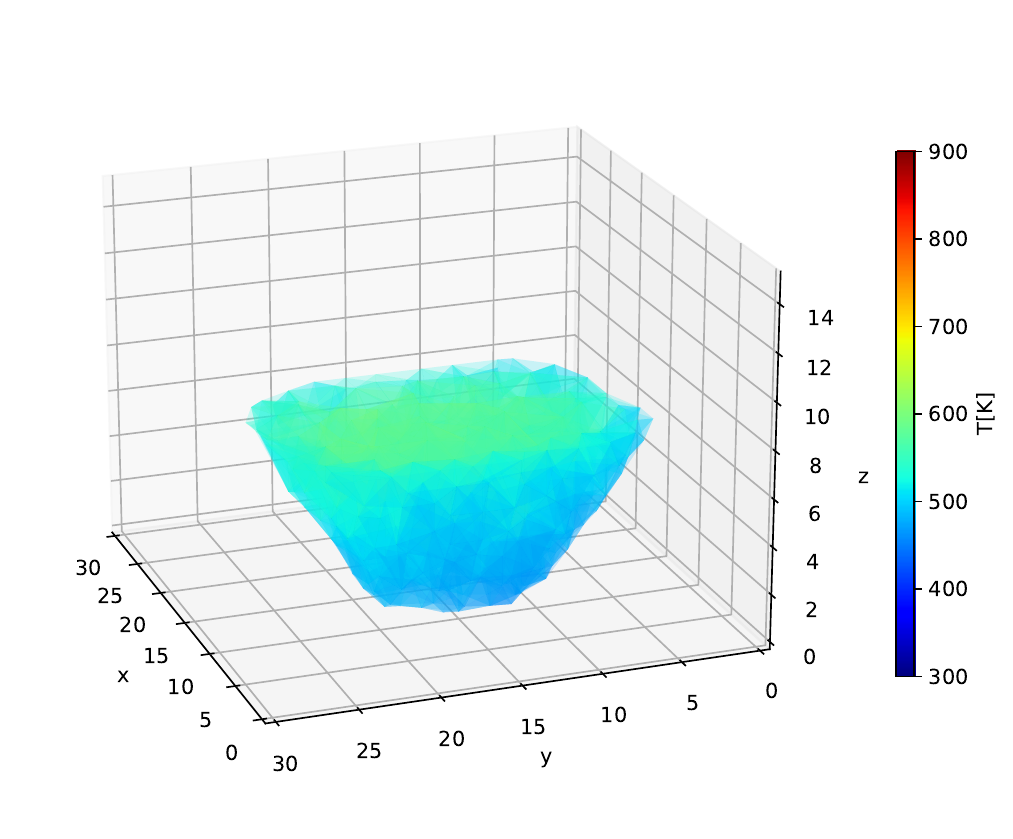}
    \includegraphics[width=.5\textwidth]{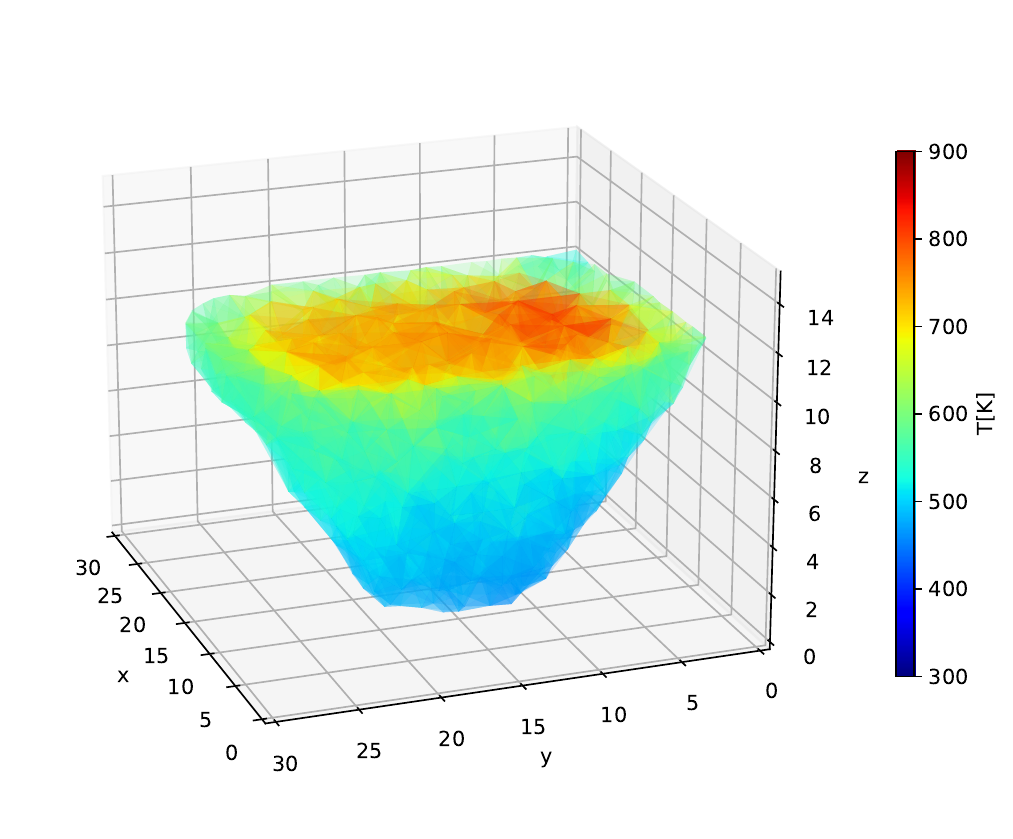}
    \includegraphics[width=.5\textwidth]{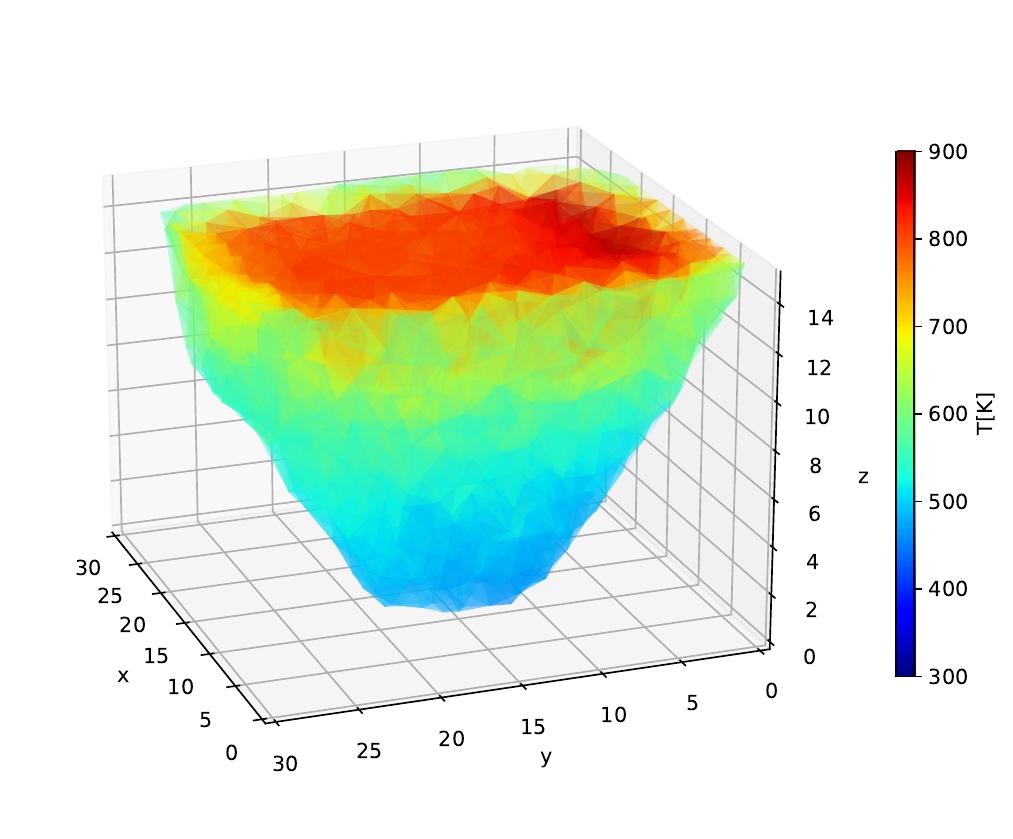}
    \caption{\textbf{4D-Heat Transport Prediction} - Heat transport evolution for layer 100, 250, 350, 500. As time progresses, the component's temperature increases, reaching a maximum at the upper portion and subsequently declining towards the base.}
    \label{fig:Heat Prediction Foundation Model}
\end{figure}

In order to predict heat transfer in PBF, the proposed model was trained for a total of 4500 iterations across 500 print layers, based on the constructed graph framework. The time steps used are measured from the beginning of the current layer to the end. The discrete heat state predictions depend on the number of thermal images, which are generated at a frequency of 3 Hz. This means that one predictive timestep corresponds to $\frac{1}{3}\, \text{s}$. The time required to print each layer varies due to the increasing surface area as the component builds up.
To enhance the discrete time resolution in each training step, we subdivide each timestep in our training data into four smaller steps for numerical approximation. This allows vertices in the lower part of the graph to influence the state at the surface. The initial state \cref{eq:initial} is taken from the first layer of the pyramid (with previous layers acting as supporting material), where the complete state is observable. This approach addresses the problem of unknown initial states. For each subsequent layer, the last predicted temperature state of the previous layer serves as the hidden initial state. The heat flux applied by the laser, modeled as boundary condition on the top surface, is represented by the Gaussian model $q= I e^{-d \eta}$, where I is the intensity, d is the distance, and $\eta$ is the decay factor. Both I and $\eta$ are data-driven parameters fitted during training. Due to the random nature of the laser trajectory, a laser detection method is employed. The Neumann boundary conditions \cref{eq: boundary_cond}, which describe radiation, are approximated using our dissipation model.
The constructed graph is illustrated in \cref{fig:temporal graph}. The temporal graph illustrates the necessary steps of the graph construction process (point cloud, simplicial complex and alpha shape) for print layers 100, 250 and 500. 
In particular, the graph undergoes substantial modifications with the addition of each layer. It is imperative to consider these changes in order to accurately capture the transient heat flow that occurs during the layer-by-layer construction in PBF. The optimisation was conducted using the ADAM optimiser. A learning rate of $\eta =\SI{1e-5}{}$ was employed, along with decay rates of $\beta_1=\num{0.5}{}$ and $\beta_2=\num{0.99}{}$, which estimate the first and second moments of the gradient to a lesser extent than is typical. The data-driven model is based on the implicit Crank-Nicolson method. \cref{fig:Heat Prediction Foundation Model} provides heat transfer snapshots at key stages of the printing process, specifically at print layers 100, 250, 350 and 500.    
This provides insight into how the model evolves to track the heat transfer dynamics at different stages. In this way, an insight can be gained into the way in which the model adapts to the ongoing thermal processes within the PBF environment. 
PiGRAND provides an intuitive visual representation of the heat transfer modeling throughout the printing process, allowing a deeper understanding of the temporal heat distribution and the influence of the increasing number of layers.
\section{Evaluation}
In this section, we conduct a comprehensive evaluation of the proposed method, focusing on prediction accuracy, performance and influence of our proposed regularization techniques.
\subsection{Print Layer Prediction Top Surface}
\begin{figure}[htbp]
\centering

\begin{minipage}{0.33\textwidth}
  \centering
  \includegraphics[width=\textwidth, height=3cm]{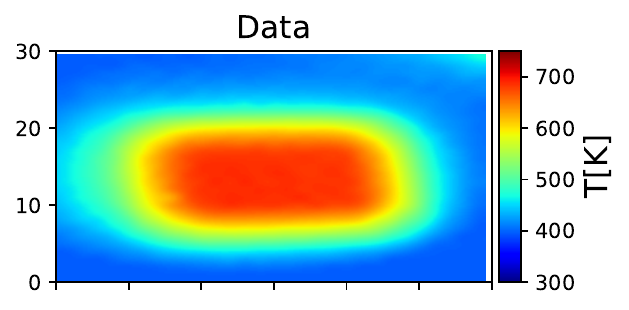} 
\end{minipage}
\hfill
\begin{minipage}{0.33333\textwidth}
  \centering
  \includegraphics[width=\textwidth, height=3cm]{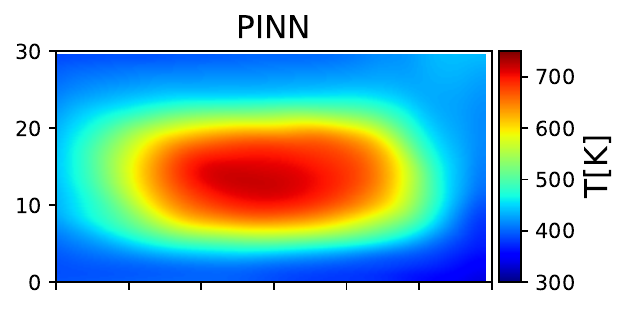} 
\end{minipage}
\hfill
\begin{minipage}{0.3\textwidth}
  \centering
  \includegraphics[width=\textwidth, height=3cm]{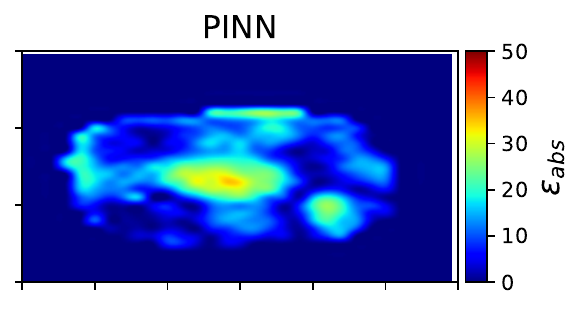} 
\end{minipage}

\vspace{0.5cm}

\hspace{0.3333333\textwidth} 
\begin{minipage}{0.33\textwidth}
  \centering
  \includegraphics[width=\textwidth, height=3cm]{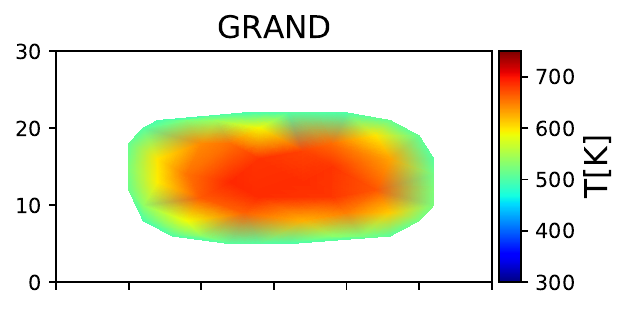} 
\end{minipage}
\hfill
\begin{minipage}{0.3\textwidth}
  \centering
  \includegraphics[width=\textwidth, height=3cm]{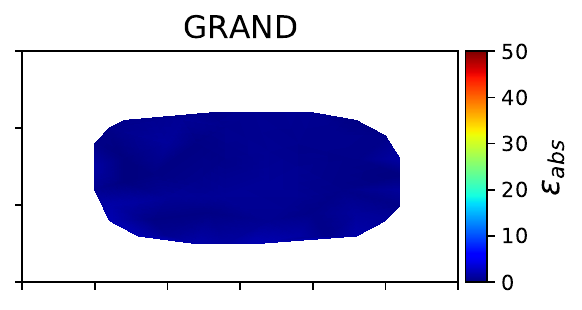} 
\end{minipage}

\vspace{0.5cm}

\hspace{0.3333333\textwidth} 
\begin{minipage}{0.33\textwidth}
  \centering
  \includegraphics[width=\textwidth, height=3cm]{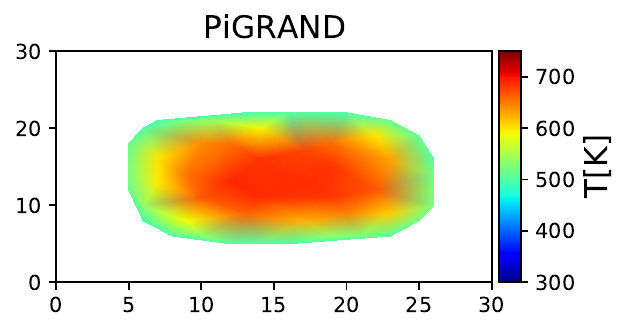} 
\end{minipage}
\hfill
\begin{minipage}{0.3\textwidth}
  \centering
  \includegraphics[width=\textwidth, height=3cm]{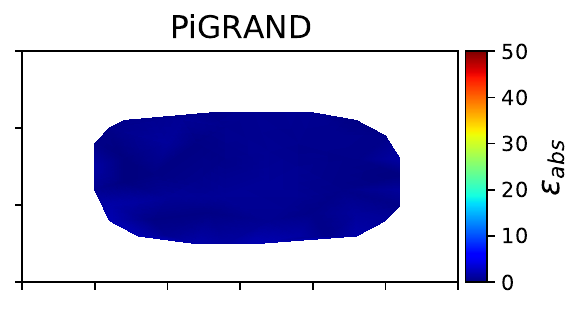} 
\end{minipage}

\caption{\textbf{Heat Transport Prediction and Error Plots} – A comparison of the heat transport prediction for print layer 232 ($t=7$s) with a PINN, GRAND, PiGRAND and the real measurement data.}
\label{fig:pred_error}
\end{figure}

 To evaluate the predictive accuracy of PiGRAND, a comprehensive comparative analysis was conducted between the observed heat state and the predicted heat state for print layer 232. These results were benchmarked against those obtained using a PINN \cite{uhrich2024physics} and traditional GRAND. Notably, GRAND predictions were based solely on the prediction loss \cref{eq: data loss} without any regularization, thereby providing a baseline for comparison (see \cref{fig:pred_error}).
\begin{figure}
    \includegraphics[width=.5\textwidth]{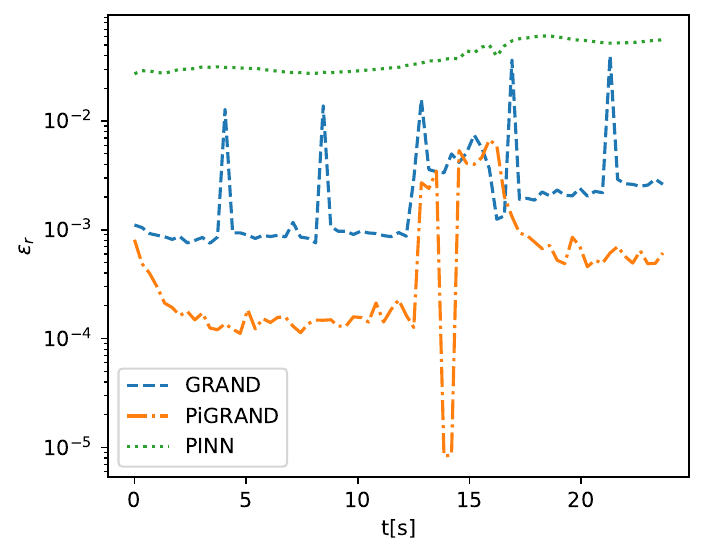}
    \caption{\textbf{Relative Error} - Comparison of $\epsilon_r$ between GRAND, PiGRAND and PINN over the entire printing layer 232 for each timepoint.}
    \label{fig:rel_error_single}
\end{figure}
In addition to the prediction of the top surface $t=7s$, the absolute error can be seen:
\begin{align}
\epsilon_{\text{abs}} = |(T_{21}(v_i)-T_{21}^{(data)}(v_i)|\,\,\, i\in top
    \label{eq:abs_error}
\end{align}
In order to evaluate the whole printing layer for all three models, we propose the relative error:
\begin{align}
\epsilon_{r} = \frac{\sqrt{\frac{1}{N}\sum_{i:\ C_i=\text{top}} \Big(T_{n+1}(v_i)-T_{n+1}^{(data)}(v_i)\Big)^2}}{\frac{1}{N}\sum_{i:\ C_i=\text{top}} \Big(T_{n+1}^{(data)}(v_i)-\frac{1}{N}\sum_{i}T_{n+1}^{(data)}(v_i)\Big)^2}
    \label{eq:rel_error}
\end{align}
\begin{table}[]
    \centering
    \begin{tabular}{|c|c|c|c|}\hline
        & PINN & GRAND & \textbf{PiGRAND}\\ 
        \hline
       $\frac{1}{N_T}\sum_{k=0}^{N_T} \epsilon_r$ &0.041 & 0.003& \textbf{0.001}\\
        \hline
    \end{tabular}
    \caption{Comparison of the mean value of $\epsilon_r$ between PINN, GRAND and PiGRAND, where $N_T$ is the number of time points during printing layer 232.}
    \label{tab:mean_rel_error}
\end{table}
The analysis included all predicted time points during the printing of layer 232, showcasing the capability of PiGRAND to accurately capture both spatial and temporal heat dynamics. The comparative results highlight PiGRAND's superior ability to handle complex heat transfer processes, demonstrating more robust and precise predictions compared to the PINN and GRAND approaches (see \cref{fig:rel_error_single} and the mean in \cref{tab:mean_rel_error}). \cref{fig:rel_error_single} illustrates the temporal evolution of the relative error $\epsilon_r$ for GRAND, PiGRAND and PINN. The results demonstrate a clear advantage of PiGRAND in terms of predictive accuracy and temporal stability.
While GRAND exhibits periodic spikes in error associated with local transient phenomena, PiGRAND significantly reduces both the magnitude and frequency of these spikes. It maintains a consistently lower error throughout the simulation highlighting its enhanced robustness and generalization.
PINNs, in contrast, exhibit higher and more stable error levels, indicating reduced sensitivity to temporal transients but also a lower overall accuracy.
\subsection{Explicit Euler vs Implicit Crank-Nicolson}
We compare the predictions of our two proposed GRAND models, inspired by the explicit Euler and the implicit Crank-Nicolson methods. It is well established in numerical analysis that implicit solvers for PDEs offer significant advantages in stability, particularly for stiff problems. However, this comes at the cost of increased computational effort. To assess the accuracy of both approaches, we employ the relative error \cref{eq:rel_error} between the data and the prediction on the top surfaces and the sum of \cref{eq:min}, \cref{eq:max}, \cref{eq:energy} and \cref{eq:heat}:
\begin{align}
    \hat{\mathcal{L}}_{\text{Energy}} = \mathcal{L}_{\text{energy}}+\mathcal{L}_{\text{heat}}+\mathcal{L}_{\text{min}}+\mathcal{L}_{\text{max}} 
\end{align}
Furthermore, to consider relevant simulation outcomes in PBF for process design, we propose analyzing the maximum temperatures of each predicted heat state. These maximum temperatures provide critical insights into the thermal behavior and are pivotal for optimizing the printing process and ensuring component quality.
\begin{figure}
    \includegraphics[width=.5\textwidth]{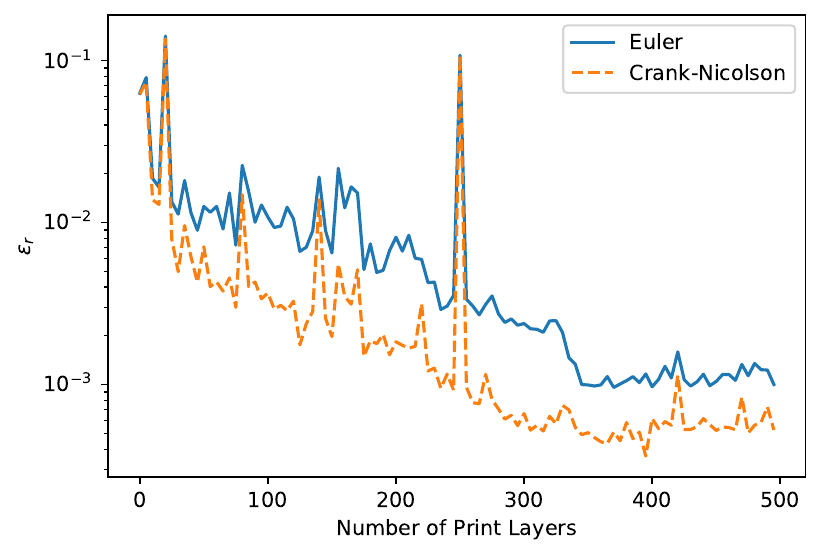}
    \includegraphics[width=.5\textwidth]{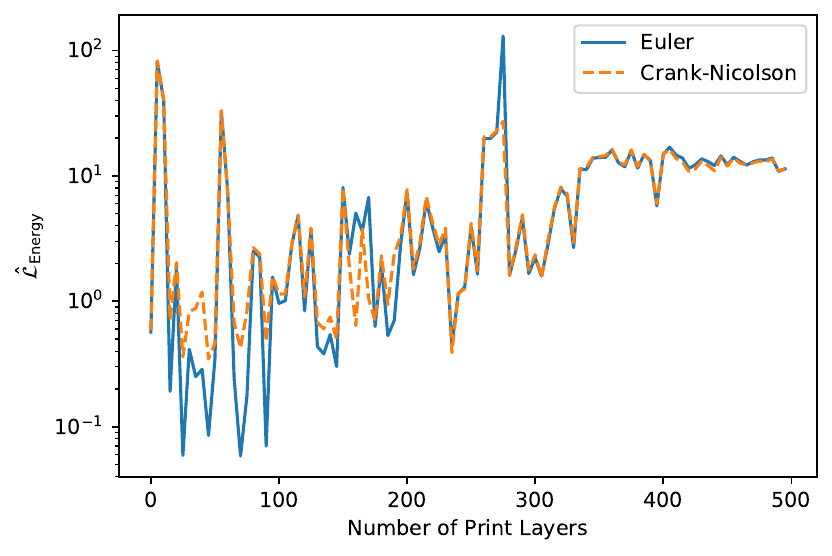}
    \includegraphics[width=.5\textwidth]{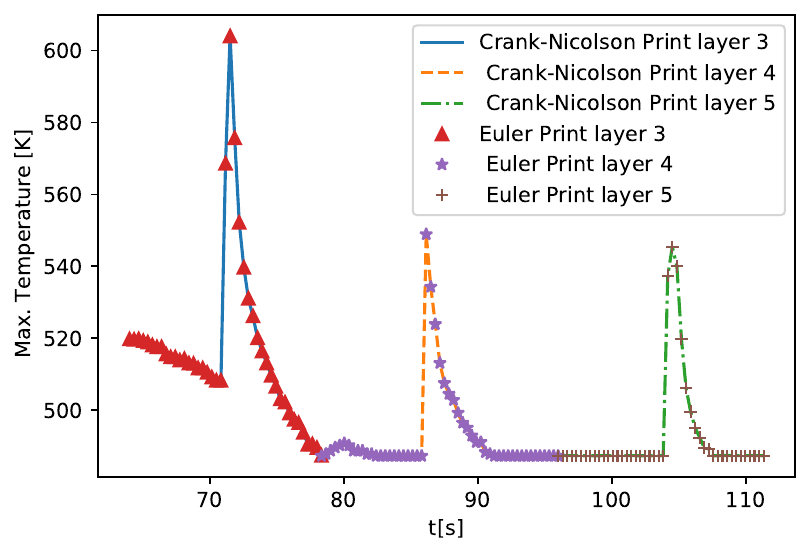}
    \caption{\textbf{Comparison of explicit Euler and implicit Crank-Nicolson inspired graph diffusion network} - Left top: Comparison of $\epsilon_r$ for 500 print layers, Right top: Comparison of $\mathcal{L}_{\text{Energy}}$ for 500 print layers. Left bottom: Max temperatures for print layers 3, 4, 5.}
    \label{fig:Explizit_Implicit}
\end{figure}
 \cref{fig:Explizit_Implicit} illustrates the accuracy of the predictions made by the Euler-inspired and Crank-Nicolson-inspired GRAND models. The Crank-Nicolson-inspired network consistently outperforms the Euler-inspired approach in terms of predictive accuracy, as evidenced in \cref{tab:Euler_Crank}.
\begin{table}[]
    \centering
    \begin{tabular}{|c|c|c|}\hline
        & $\frac{1}{N_L}\sum_{k=0}^{N_L} \epsilon_r$ & $\frac{1}{N_L}\sum_{k=0}^{N_L} \hat{\mathcal{L}}_{\text{Energy}}$\\ 
        \hline
        Euler & 0.010 & 24.473\\
        \hline
         Crank-Nicolson & \textbf{0.006} & \textbf{23.449}\\
        \hline
    \end{tabular}
    \caption{Comparison of the mean of $\epsilon_r$ and $\mathcal{L}_{Energy}$ between Euler and Crank-Nicolson inspired Graph Neural Diffusion, where $N_L$ is the number of print layers.}
    \label{tab:Euler_Crank}
\end{table}
This superior performance can be attributed to the enhanced stability and numerical robustness inherent to the Crank-Nicolson method. The energy loss $\hat{\mathcal{L}}_{Energy}$ values reported in \cref{tab:Euler_Crank} and  \cref{fig:Explizit_Implicit} are indeed higher than the relative error, but this is expected due to the way the loss is structured. The total energy loss $\hat{\mathcal{L}}_{Energy}$ is a composite loss that includes multiple terms. These terms capture different aspects of the thermal field, such as conservation of energy, heating dynamics and extreme temperatures. Since the loss accumulates several components, its absolute scale is naturally larger than individual metrics like the relative error $\epsilon_r$. Importantly, the loss shows a consistent convergence trend across print layers and the models prediction remain physically plausible and accurate.
 Based on these findings, we adopt the Crank-Nicolson-inspired approach for all subsequent evaluations.
\subsection{Regularization}
\begin{table}[]
    \centering
    \begin{tabular}{|c c|c|c|c|c|}\hline
        regularization & weight & $\mathcal{L}_\text{data}$ & $\mathcal{L}_{\varphi}$ & $\mathcal{L}_{\psi}$ & $\mathcal{L}_\text{energy}$\\\hline
         & high &  \textbf{1.085} & \textbf{1.134} & 0.375 & \textbf{0.000}\\
        all & normal & 1.128 &  1.211 & \textbf{0.175} & 0.076 \\
         & low & 1.111 &  1.206 & 0.231 &0.588\\\hline
        & high & 1.107 & 1.205 & 0.238 & 0.367 \\
        $\mathcal{L}_\varphi,\, \mathcal{L}_\psi,\, \mathcal{L}_\text{heat}$ & normal & 1.109 & 1.205 & 0.238 & 0.335 \\
        & low & 1.108 & 1.205 & 0.269 & 0.667\\ \hline
         & high & 1.100 & 1.319 & 0.251 & \textbf{0.000}\\
        $\mathcal{L}_\text{min},\mathcal{L}_\text{max},\mathcal{L}_\text{energy}$ & normal & 1.140 & 1.211 & 0.223 & 0.418\\
        & low & 1.135 & 1.207 & 0.274  & 1.389\\ \hline
        & none & 1.133 & 1.206 & 0.313 & 1.593\\
        \hline
    \end{tabular}
    \caption{Model fit for the foundation model using different sets of weighted regularization functions for training. 'High weights' uses a factor of $10$ for the regularization terms in the total loss function, 'low weights' uses a factor of $0.1$.}
    \label{tab: regularization}
\end{table}
A comprehensive study on the influence of incorporating the proposed physical principles and mathematical constraints into the loss function is presented in \cref{tab: regularization} and \cref{tab: regularization_pyramid_4}, for the models of P7 and P4, respectively. We investigate four training configurations:
\begin{itemize}
\item All regularization losses are included: $\mathcal{L}_{\text{data}}$, $\mathcal{L}_{\varphi}$, $\mathcal{L}_{\psi}$, $\mathcal{L}_{\text{energy}}$, $\mathcal{L}_{\text{heat}}$, $\mathcal{L}_{\text{min}}$ and $\mathcal{L}_{\text{max}}$.
\item Only mathematically derived constraints are used: $\mathcal{L}_{\varphi}$, $\mathcal{L}_{\psi}$, and $\mathcal{L}_{\text{heat}}$.
\item Only physics-based constraints are used: $\mathcal{L}_{\text{min}}$, $\mathcal{L}_{\text{max}}$ and $\mathcal{L}_{\text{energy}}$.
\item No regularization only data fitting via $\mathcal{L}_\text{data}$.
\end{itemize}
Each trained model is evaluated based on four loss components: $\mathcal{L}_{\text{data}}$, $\mathcal{L}_{\varphi}$, $\mathcal{L}_{\psi}$, and $\mathcal{L}_{\text{energy}}$.
To assess the sensitivity of the model to regularization strength, we define three weighting levels for the regularization losses relative to the data loss: high (10×), normal (1×) and low (0.1×). These weights are chosen heuristically and calibrated through preliminary convergence experiments to reflect varying emphasis on physical consistency during training.
To systematically explore the role of regularization, an ablation study is conducted across the three weighting scenarios. As shown in \cref{tab: regularization} and \cref{tab: regularization_pyramid_4}, increasing the weight of the physics-based loss terms generally leads to a reduction in $\mathcal{L}_{\text{energy}}$, reflecting improved physical consistency. Notably, this improvement is achieved without significantly compromising the data fit, as indicated by stable values for $\mathcal{L}_{\text{data}}$.
To ensure a balanced contribution of all loss terms to the overall objective, each individual loss is normalized during training, accounting for potential differences in magnitude.
\begin{table}[]
    \centering
    \begin{tabular}{|c c|c|c|c|c|}\hline
        regularization & weight & $\mathcal{L}_\text{data}$ & $\mathcal{L}_{\varphi}$ & $\mathcal{L}_{\psi}$ & $\mathcal{L}_\text{energy}$\\\hline
         & high & 1.067 &  1.021 & 0.133 & \textbf{0.000}\\
        all & normal & 1.065 &  \textbf{0.979} & 0.127 & 0.008 \\
         & low & 1.063 &  0.981 & 0.114 &0.291\\\hline
        &high & 1.063 & \textbf{0.979} & 0.112 & 0.029 \\
        $\mathcal{L}_\varphi,\, \mathcal{L}_\psi,\, \mathcal{L}_\text{heat}$ & normal & 1.063 & \textbf{0.979} & 0.108 & 0.024 \\
        & low & 1.063 & 0.981& 0.104 & 0.283\\ \hline
         & high & 1.077 & 1.010 & \textbf{0.101} & \textbf{0.000}\\
        $\mathcal{L}_\text{min},\mathcal{L}_\text{max},\mathcal{L}_\text{energy}$ & normal & 1.064 & 0.985 & 0.102 & 0.955\\
        & low & \textbf{1.062} & 0.985 & 0.111  & 1.490\\ \hline
        & none & \textbf{1.062} & 0.986 & 0.118 & 1.556\\
        \hline
    \end{tabular}
    \caption{Model fit for P4 using different sets of weighted regularization functions for training. 'High weights' uses a factor of $10$ for the regularization terms in the total loss function, 'low weights' uses a factor of $0.1$.}
    \label{tab: regularization_pyramid_4}
\end{table}
For the fit of the foundation model (see \cref{tab: regularization}), assigning high weights to all regularization losses yields the best overall results. In particular, optimal performance is achieved for $\mathcal{L}_{\text{data}}$, $\mathcal{L}{\varphi}$, and $\mathcal{L}_{\text{energy}}$. In contrast, a GRAND model trained without any regularization performs poorly across all four metrics, especially with regard to $\mathcal{L}_{\text{energy}}$.
It is important to note that the impact of individual losses such as $\mathcal{L}_{\text{data}}$, $\mathcal{L}_{\varphi}$, and $\mathcal{L}_{\psi}$ is relatively minor. This supports the view that GRAND is effective in data assimilation, but lacks the capability to predict the temperature distribution within the component, particularly in regions where no sensor data is available.
A similar observation can be made in the study for the fit of P4 (see \cref{tab: regularization_pyramid_4}). Unlike the foundation model, a clear trade-off between $\mathcal{L}_{\text{energy}}$ and $\mathcal{L}_{\text{data}}$ is evident. The best data loss is achieved without any regularization, but this leads to the worst performance in energy consistency. Thus, GRAND without regularization is not suitable when physical law violations are critical.
From both experiments, we conclude that , training with all regularization terms delivers the most balanced performance.
However, training with only physical principles (excluding $\mathcal{L}_{\varphi}, \mathcal{L}_{\psi}, \mathcal{L}_{heat}$) also provides good results, especially for reducing energy violations without turning the training into a highly multi-objective problem.
In summary, while traditional GRAND is capable of learning surface-level heat transfer patterns, it fails to generalize to the internal temperature distribution of the component, particularly in sensor-sparse areas. Our proposed use of physical principles as regularizers offers a significant benefit in prediction accuracy and physical consistency.
The final solution consists of a hybrid digital twin, capable of digitally replicating both the physical structure of the printed component and the dynamic heat transfer process occurring within it.
\subsection{Inference and Transfer Learning based on Foundation Model}
\begin{figure}
    \includegraphics[width=.5\textwidth]{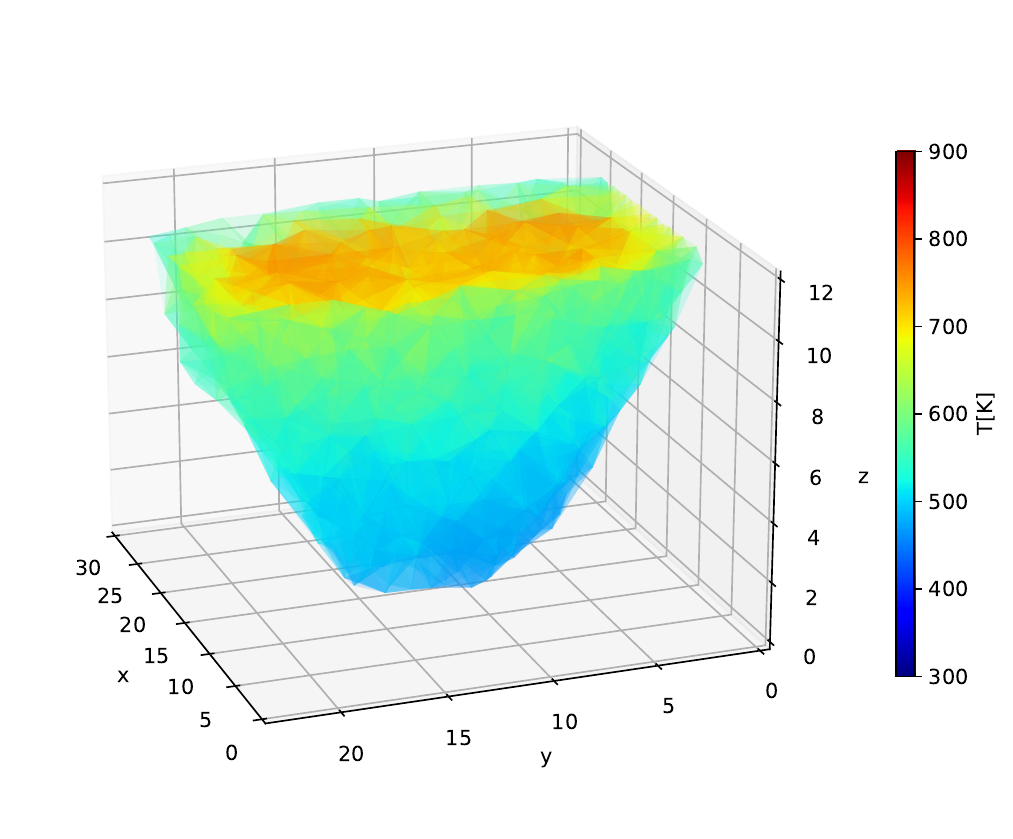}
    \includegraphics[width=.5\textwidth]{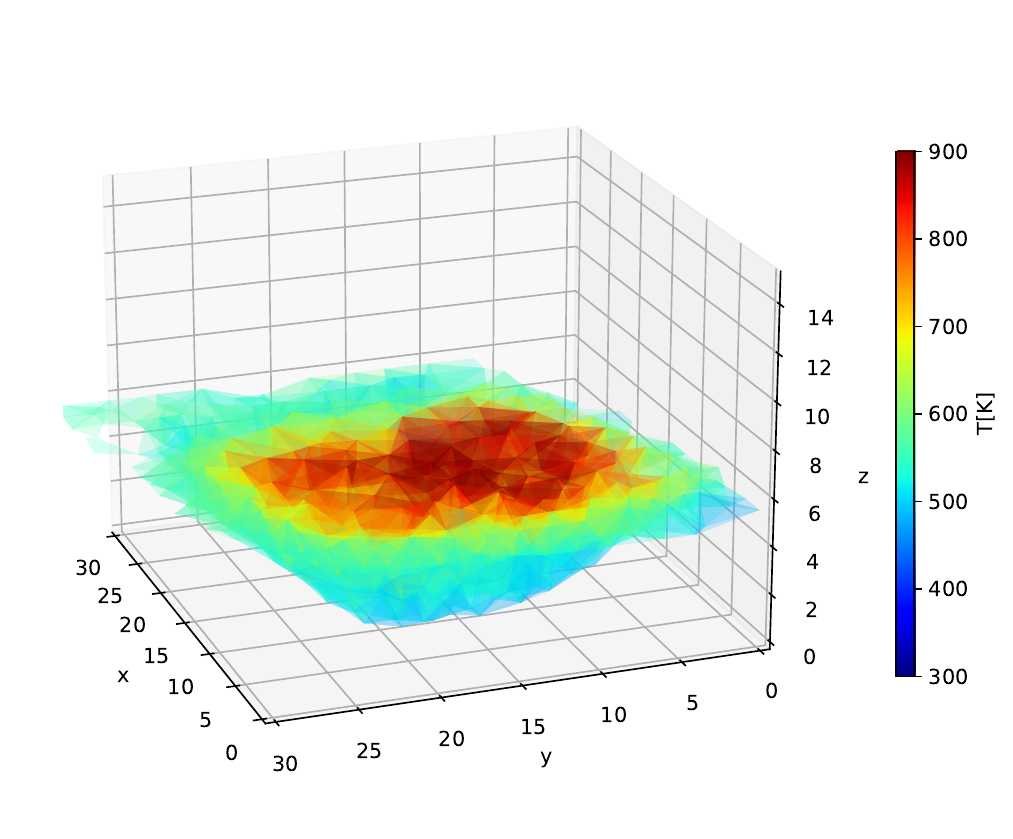}
     \includegraphics[width=.5\textwidth]{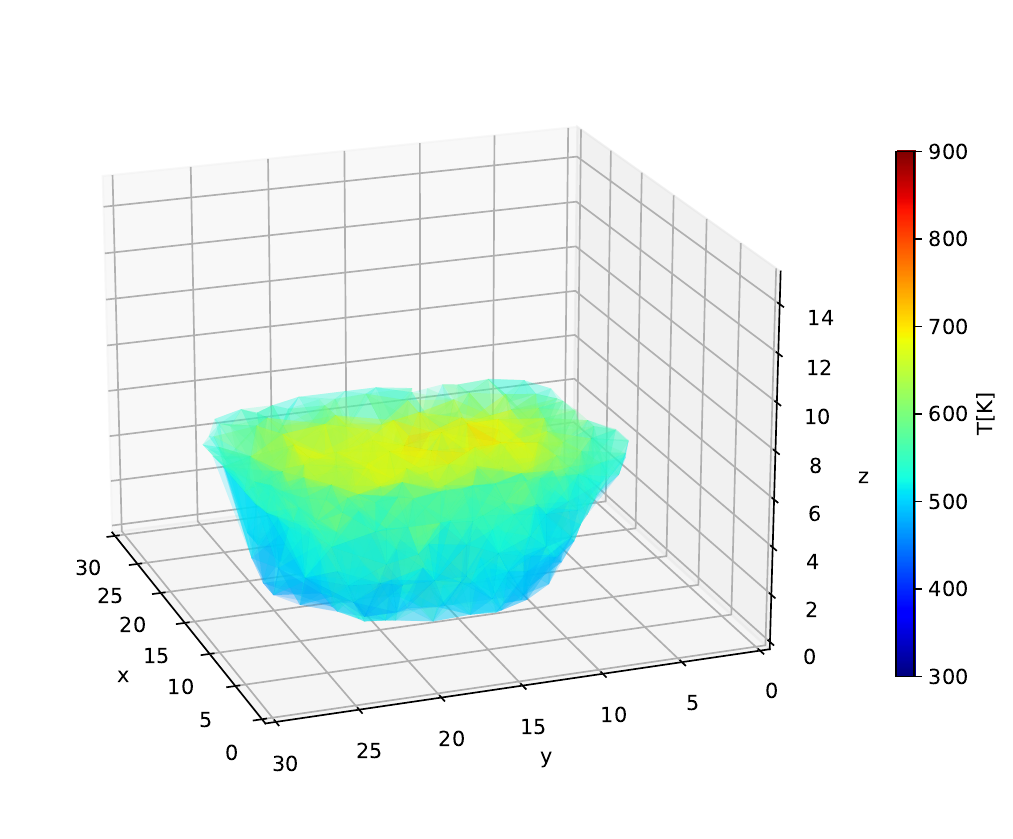}
    \caption{\textbf{Inference and transfer learning} - Left: Inference for pyramid P4, print layer 380, Right: Inference for pyramid P9, print layer 200, Bottom: Transfer learned heat transport for pyramid P7M, print layer 232.}
    \label{fig:Tranfer_Prediction}
\end{figure}

To reduce computational effort, we propose an efficient transfer learning strategy. Traditional training from scratch for each new model is computationally expensive, particularly when dealing with components of similar geometry but differing in size or material. Instead, we utilize the previously discussed foundation model, trained on stainless steel, as a pretrained base.  This allows us to transfer its learned thermal behaviour to new components, such as those made from nickel-based alloys. Moreover, the same pretrained model can be directly used for inference on components made from the same material but with varying sizes. This is feasible because the temperature transport dynamics for the material have already been effectively learned. As illustrated in \cref{fig:Tranfer_Prediction}, we apply the pretrained model to predict the thermal behaviour of high-quality components P4 (stainless steel) and P7M (nickel-based alloy). Additionally, we analyse P9 (stainless steel), which exhibits low quality and significant structural deformation (see \cref{fig:real-components}). In particular, if P9 had been manufactured to a higher standard, the maximum temperature would likely be reduced, and the final build height would be greater. This shows that the model not only predicts thermal distributions, but also provides insight into process anomalies, such as structural defects.\\
We acknowledge the risk of overfitting or bias when evaluating the model on data it has already seen. To ensure proper generalization, we evaluate model predictions on unseen data. Importantly, data for P4 and P9 were excluded from the training phase, providing a robust assessment of the model’s ability to generalize. The results confirm that the model achieves high accuracy on previously unseen components, demonstrating its transferability and reliability.\\     
\begin{figure}
    \includegraphics[width=.5\textwidth]{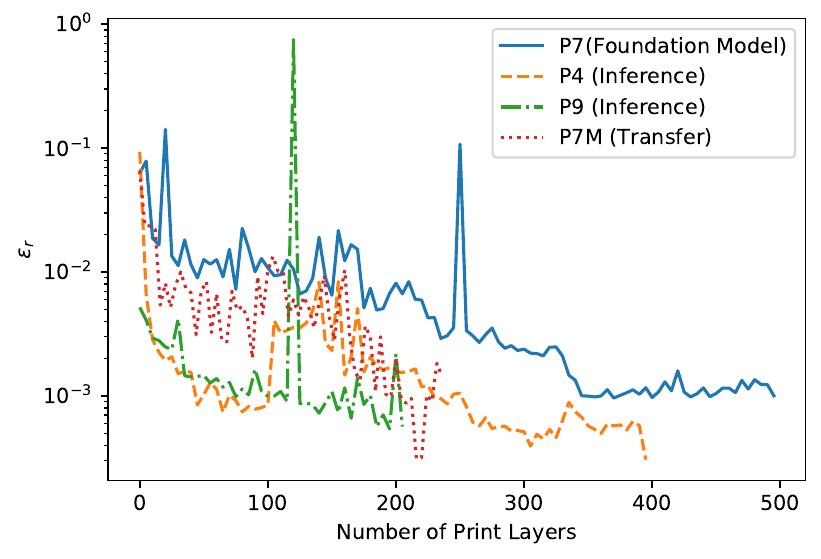}
    \includegraphics[width=.5\textwidth]{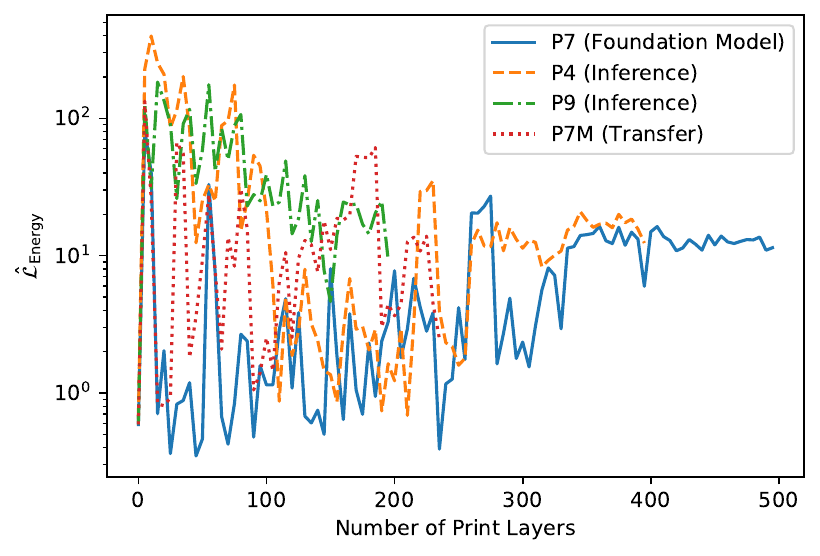}
    \caption{\textbf{A comparison of the P7 foundation model inspired by Crank-Nicolson, the P4 and P9 inference and the P7M transfer-learned model.} - Left: Comparison of $\epsilon_r$ for 500 print layers, Right: Comparison of $\mathcal{L}_{\text{Energy}}$ for 500 print layers.}
    \label{fig: Evaluation_transfer}
\end{figure}
In addition to relative error metrics, we employ energy loss ($\mathcal{L}_{\text{Energy}}$) as a complementary evaluation criterion. \cref{fig: Evaluation_transfer} presents the relative error $\epsilon_r$ and energy loss $\mathcal{L}_{\text{Energy}}$ across the entire set of printed layers, including the foundation model (P7), the inference to P9 and P4 as well as the transfer learned model of P7M. The low error values validate the effectiveness of the transfer-learned model in capturing the key thermal behaviors.
Finally, conventional 3D printing simulations require re-discretization and re-solving of the PDEs for each new part. This is the case even for minor changes in geometry. Our approach mitigates this inefficiency. During the initial training phase, the model learns the underlying physics, enabling rapid and physically consistent inference and significantly improving computational efficiency.
\subsection{Performance}
\begin{figure}
    \includegraphics[width=.5\textwidth]{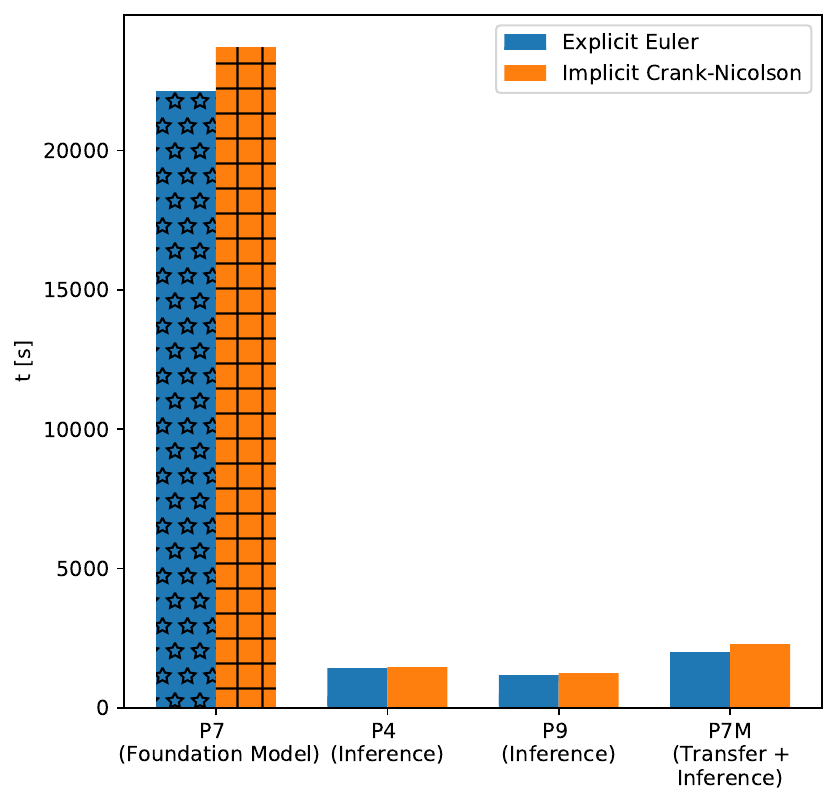}
    \includegraphics[width=.5\textwidth]{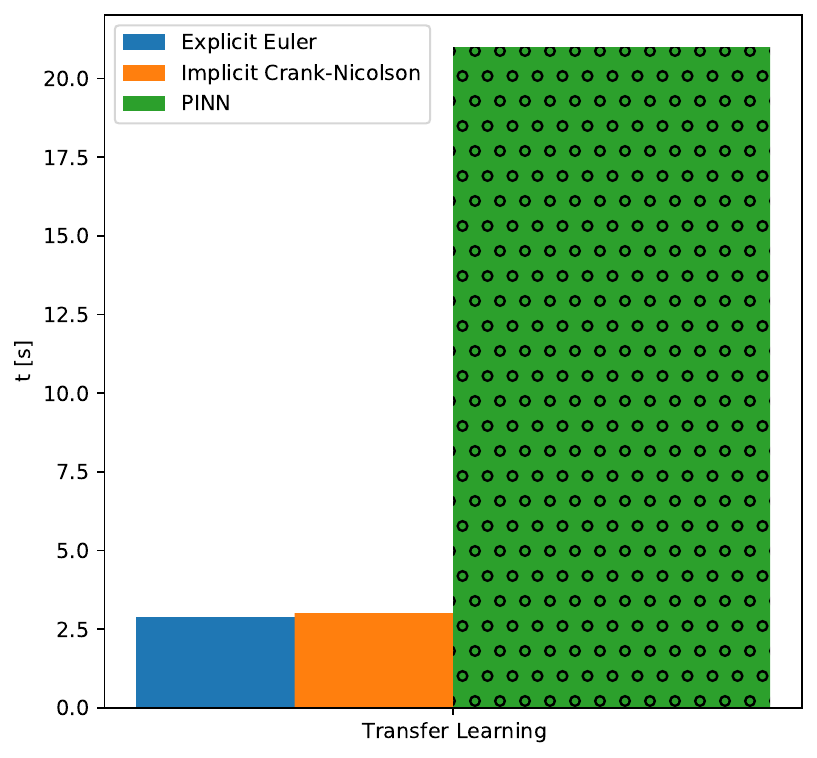}
    \includegraphics[width=.5\textwidth]{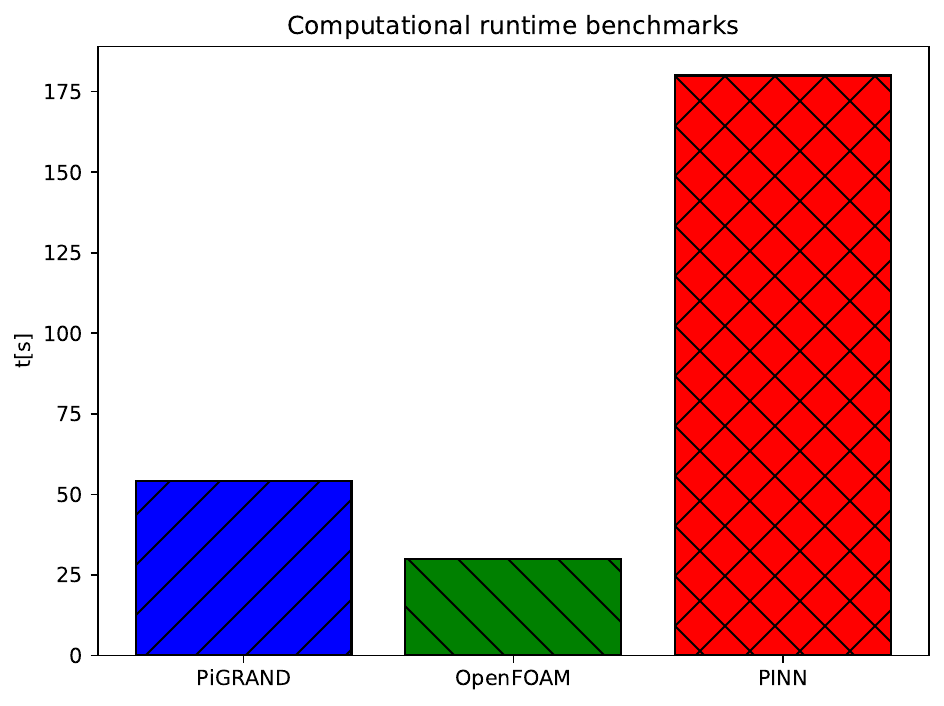}
     \caption{\textbf{Performance of PiGRAND} - Left: The total solving time of the PiGRAND foundation model is compared with the inference times of the other printed components P4, P9 and the transfer-learned temperature distribution of P7M, Right: The average computational effort associated with the prediction of the temperature distribution of a single printed layer, contrasting the performance of a transfer-learned PINN with our proposed PiGRAND, Bottom: The total solving time of PiGRAND (Crank-Nicolson) compared to a FVM solver and a vanilla PINN for one printing benchmark layer}
    \label{fig: Performance}
\end{figure}

In addition to evaluating the quality and capabilities of the predictions, we assessed the computational effort required by the proposed models. Specifically, we investigated the training time required for the explicit and implicit foundation models inspired by the Euler and Crank–Nicolson methods, respectively. The solving time of the transfer-learned model was compared to this. We also compared it to the inference prediction of P4, P9 and P7M. Furthermore, we evaluated the average computational solving time of PiGRAND for a printed layer against that of a transfer-learned PINN from \cite{uhrich2024physics}, as shown in \cref{fig: Performance}.

As expected, the computational times for the Crank–Nicolson-inspired models were slightly higher than those for the explicit Euler-inspired models, due to the increased stability and complexity of the implicit method. However, once the foundation model has been trained, the computational effort required to predict the temperature distribution of components with similar geometries is significantly reduced. Rather than repeatedly solving the diffusion equation for each new geometry, PiGRAND enables real-time or near-real-time predictions, requiring considerably less computational effort.
Predictions can be made for new components and the same material with a single forward pass through the network. This approach incurs a significantly lower computational cost than training the model from scratch. PiGRAND can predict the heat transport of new components made of the same material 15 times faster than the initial predictions made by the foundation models. Additionally, PiGRAND can predict a printed layer seven times faster than the transfer-learned PINN model presented in \cite{uhrich2024physics}.  For components made of different materials, the pretrained foundation model can be efficiently retrained and fine-tuned.  
To substantiate the efficiency of PiGRAND, we conducted a comparative runtime benchmark against a traditional finite volume solver (OpenFOAM) and PINN for a benchmark problem presented in \cite{uhrich2024physics}. PiGRAND significantly outperforms PINNs in terms of computational efficiency, requiring only 54.27 sec, compared to over 175 seconds for the PINN approach. While OpenFOAM remains faster in this benchmark, PiGRAND offers a compelling trade-off by combining data efficiency and generalization capabilities of GNNs with substantially lower computational cost than standard PINNs.
These results demonstrate that PiGRAND achieves a 3–4× speedup over PINNs while preserving physical interpretability and accuracy, highlighting its promise for practical deployment in computationally constrained environments.
\cref{fig: Performance} shows the time required to complete inference on the same problem setup.
Our results demonstrate that PiGRAND outperforms PINNs in both prediction accuracy and computational efficiency, highlighting its superiority for heat transport prediction in 3D printing.      
\section{Conclusion}
In this work, we proposed PiGRAND for predicting heat transfer in 3D-printed components. Building upon the model ideas of GRAND by Chamberlain \textit{et al.}, we accelerated the learning process through the integration of mathematical regularization principles and physical constraints derived from the theoretical study of PDEs. Additionally, we introduced a novel connectivity model and classifier method, enabling nodes and edges to acquire distinct spatial characteristics. Thermal images representing real measurement data were incorporated into the model through a generated graph data structure. To address energy transfer at the boundaries, we developed an dissipation model that learns these dynamics effectively.\\
We utilized both explicit Euler and implicit Crank-Nicolson-inspired graph neural diffusion approaches and evaluated their prediction accuracy. The superior performance of the Crank-Nicolson-inspired model led us to further investigate the impact of our proposed regularizers compared to GRAND. Acknowledging the computational intensity of training such models, we presented an efficient transfer learning strategy that leverages a pretrained foundation model to predict heat transfer in components with similar geometries but different materials. Our evaluation demonstrates the substantial computational efficiency of our approach, outperforming traditional methods and achieving significantly faster predictions. Additionally, PiGRAND was benchmarked against a PINN, with results showcasing better accuracy and faster computational solving times.\\
Over the last decades, numerical analysis methods like FEM and FVM have achieved significant milestones in simulating thermal processes. However, these methods exhibit limitations compared to PiGRAND. For instance, numerical approaches require domain experts to discretize components into a mesh or control volumes, a process heavily dependent on geometry and mesh resolution. In contrast, our data-driven graph construction offers greater flexibility and automation, eliminating the need for domain knowledge. Furthermore, our connectivity model enables discretization, where nodes play distinct roles and hold unique properties in space. While FEM and FVM are limited in their data-driven capabilities, they also rely on well-posed physical problems with precise boundary conditions. PiGRAND, however, incorporates an intelligent dissipation model.
We demonstrated the benefits of our approach over state-of-the-art machine learning models. Although this work provides a strong proof of concept with promising results, there are several opportunities for future research and refinement. The connectivity function $c_{ij}$ and the dissipation function $Q_i$ are modelled as single-hidden-layer neural networks with width 256. While one hidden layer has theoretical universality, the required width can become exponentially large for complex high-dimensional functions \cite{cybenko1989approximation, hornik1991approximation}. This makes shallow networks inefficient compared to deeper ones. If the mapping were more complex or highly nonlinear, deeper architectures can achieve similar approximation accuracy with far fewer neurons (see e.g. \cite{eldan2016power, telgarsky2016benefits}. This might be considered for future works. Our approach should be evaluated on more complex geometries to test its generalization capabilities. Furthermore, extending PiGRAND to address a broader range of physical problems and PDEs could solidify its applicability beyond 3D printing. Hyperparameter optimization was not performed in this work and there is potential for further performance improvements in this regard. Lastly, while parallelization offers an avenue for accelerating computational performance, implementing it in this context poses challenges due to the sequential dependency of heat state predictions.
\section*{Acknowledgement}
The authors acknowledge the assistance and resources provided by the SIEMENS AG, which were instrumental in the data generation efforts for this project. This work was supported by Martin Schäfer and Oliver Theile, enabling access to the thermal images that document the heat transfer on the surface of the components during 3D printing.
The authors acknowledge the financial support by the Federal Ministry of Education and Research of Germany and by the S\"achsische Staatsministerium f\"ur Wissenschaft Kultur und Tourismus in the program Center of Excellence for AI-research "Center for Scalable Data Analytics and Artificial Intelligence Dresden/Leipzig", project identification number: ScaDS.AI.
\section*{Statements and Declarations}
\subsection*{Funding}
The work is funded by the the Federal Ministry of Education and Research of Germany and by the S\"achsische Staatsministerium f\"ur Wissenschaft Kultur und Tourismus in the program Center of Excellence for AI-research "Center for Scalable Data Analytics and Artificial Intelligence Dresden/Leipzig", project identification number: ScaDS.AI
\subsection*{Competing Interests}
The authors do not have any relevant financial or non-financial interests to report.
The authors have no competing interests to declare that would be relevant to the content of this article.
All authors certify that they have no affiliation or involvement with any organisation or entity that has any financial or non-financial interest in the subject matter or materials discussed in this manuscript.
No material discussed in this article has any financial or proprietary interest for the authors.
\subsection*{Authors Contribution}
Conceptualization: Benjamin Uhrich, Methodology: Benjamin Uhrich, Tim Häntschel, Formal analysis and investigation: Benjamin Uhrich, Tim Häntschel; Visualization: Benjamin Uhrich, Tim Häntschel; Writing - original draft preparation: Benjamin Uhrich, Tim Häntschel; Writing - review and editing: Benjamin Uhrich, Tim Häntschel, Erhard Rahm; Supervision: Erhard Rahm
\subsection*{Code and Data Availability}
All the code is open sourced and available on GitHub: https://github.com/bu32loxa/PiGRAND
\bibliography{references.bib}

\end{document}